\documentclass[10pt,twocolumn,letterpaper]{article}

\usepackage[pagenumbers]{iccv} 
\usepackage{graphicx}
\usepackage{amsmath}
\usepackage{amssymb}
\usepackage{booktabs}
\usepackage{animate}
\usepackage{enumitem}
\usepackage{multirow}
\usepackage{diagbox}

\usepackage{graphicx}
\usepackage{amsmath}
\usepackage{amssymb}
\usepackage{booktabs}
\usepackage{float}
\usepackage{diagbox}
\usepackage{multirow}
\usepackage{enumitem}
\usepackage{tabularx}
\usepackage{microtype}
\usepackage{soul}

\usepackage[numbers,sort&compress]{natbib}

\usepackage{adjustbox} 

\makeatletter
\@namedef{ver@everyshi.sty}{}
\makeatother
\usepackage{tikz}
\usepackage{makecell}
\usepackage{pgfplots}
\usetikzlibrary{spy,calc}
\definecolor{iccvblue}{rgb}{0.21,0.49,0.74}
\usepackage[pagebackref,breaklinks,colorlinks,allcolors=iccvblue]{hyperref}

\usepackage[capitalize]{cleveref}
\crefname{section}{Sec.}{Secs.}
\Crefname{section}{Section}{Sections}
\Crefname{table}{Table}{Tables}
\crefname{table}{Tab.}{Tabs.}

\newcommand{\rev}[1]{\textcolor{black}{#1}}  

\newcommand{\revision}[1]{\textcolor{black}{#1}}

\def\paperID{2976} 
\def\confName{ICCV}
\def\confYear{2025}

\definecolor{Gray}{gray}{0.5}
\definecolor{LightCyan}{rgb}{0.88,1,1}

\newcolumntype{a}{>{\columncolor{Gray}}c}
\newcolumntype{b}{>{\columncolor{white}}c}

\begin{document}


\title{Stable Score Distillation} 

\author{Haiming Zhu$^1$,
       Yangyang Xu$^2$,
       Chenshu Xu$^1$,
       Tingrui Shen$^3$,\\
       Wenxi Liu$^4$,
       Yong Du$^5$,
       Jun Yu$^2$,
       Shengfeng He$^1$
       \\
       \normalsize{$^1$Singapore Management University \hspace{5mm} $^2$Harbin Institute of Technology (Shenzhen)} \\
       \normalsize{$^3$South China University of Technology \hspace{5mm} $^4$Fuzhou University \hspace{5mm} $^5$Ocean University of China \vspace{-1mm}}}

\teaser{
    \centering
    \begin{subfigure}{.18\linewidth}
        \centering
        \includegraphics[width=\textwidth]{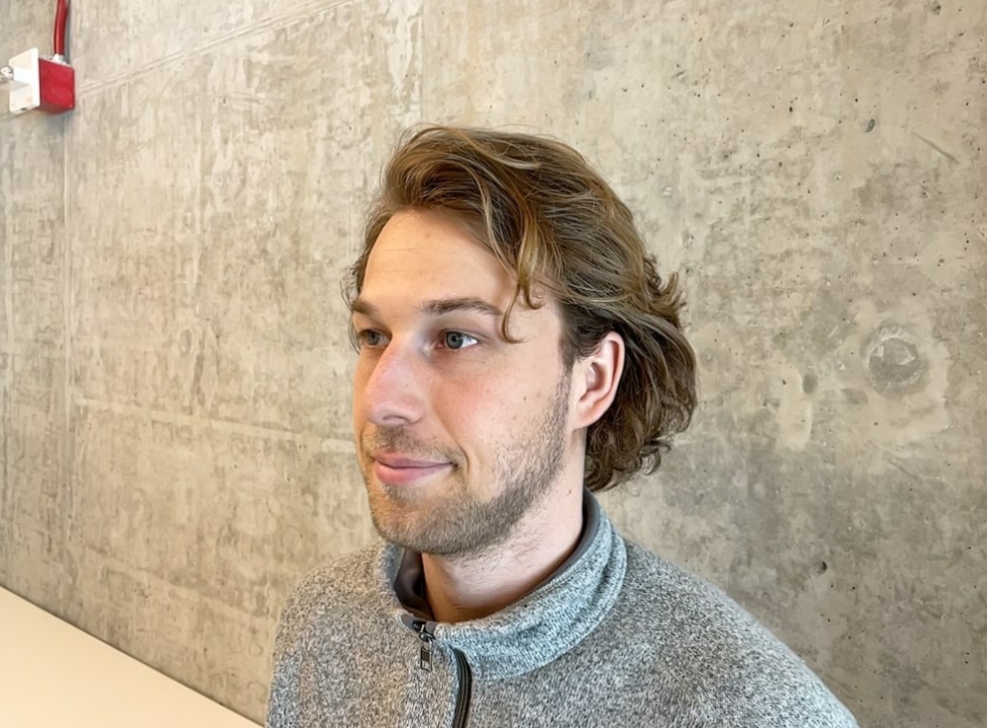}\\
        \includegraphics[width=\textwidth]{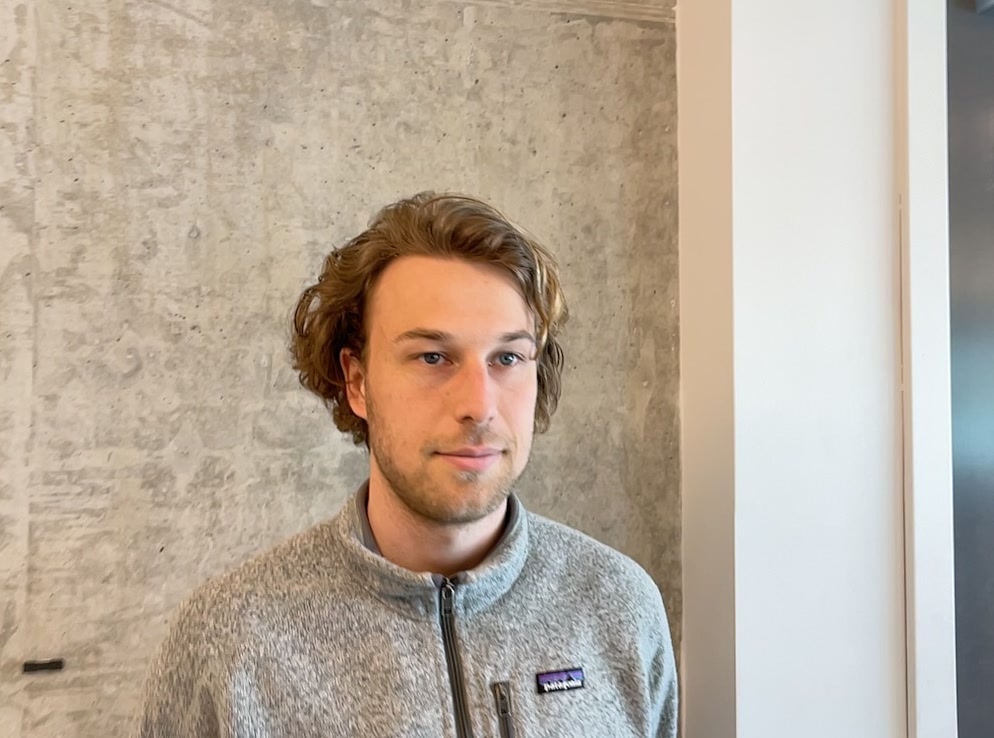}
        \caption{\footnotesize{Original Views}}
        \label{fig:fig_original}
    \end{subfigure}
    \begin{subfigure}{.265\linewidth}
        \centering
        \begin{minipage}{\textwidth}
            \centering
            \includegraphics[width=\textwidth]{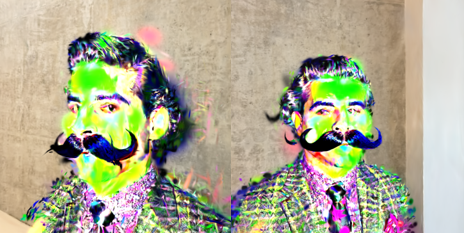}
        \end{minipage}
        \begin{minipage}{\textwidth}
            \centering
            \includegraphics[width=\textwidth]{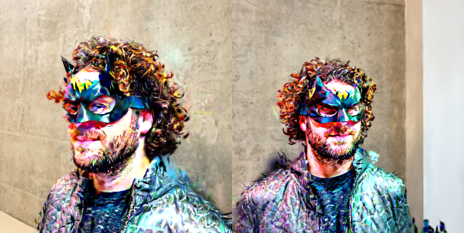}
        \end{minipage}
        \caption{\footnotesize{{CSD\cite{yu2024texttod}}}}
        \label{teaser:csd}
    \end{subfigure}
    \hspace{-1.5mm}
    \begin{subfigure}{.265\linewidth}
        \centering
        \begin{minipage}{\textwidth}
            \centering
            \includegraphics[width=\textwidth]{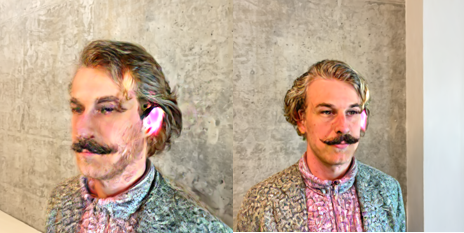}
        \end{minipage}
        \begin{minipage}{\textwidth}
            \centering
            \includegraphics[width=\textwidth]{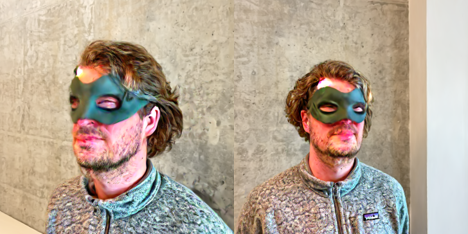}
        \end{minipage}
        \caption{\footnotesize{{DDS\cite{hertz2023delta}}}}
        \label{teaser:dds}
    \end{subfigure}
    \hspace{-1.5mm}
    \begin{subfigure}{.265\linewidth}
        \centering
        \begin{minipage}{\textwidth}
            \centering
            \includegraphics[width=\textwidth]{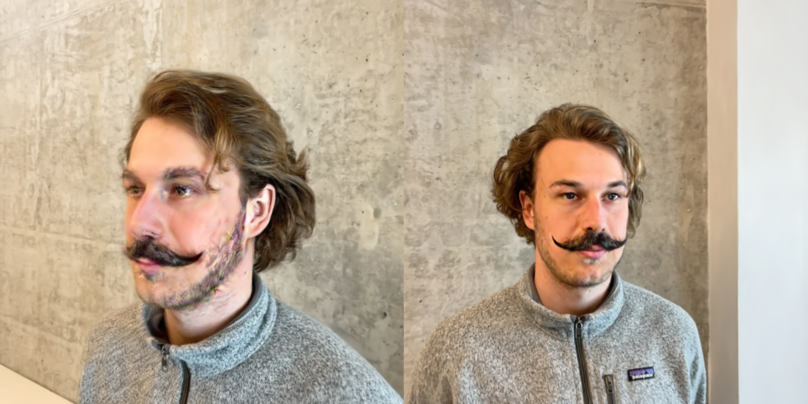}
        \end{minipage}
        \begin{minipage}{\textwidth}
            \centering
            \includegraphics[width=\textwidth]{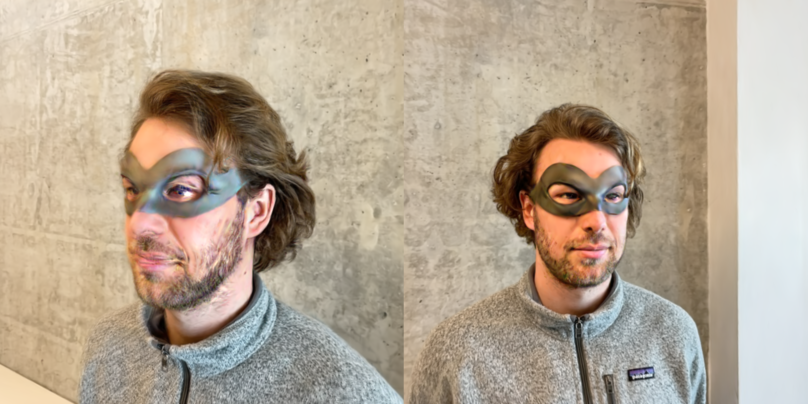}
        \end{minipage}
        \caption{\footnotesize{{Ours}}}
        \label{teaser:ours}
    \end{subfigure}
    \begin{minipage}{.001\linewidth}
    \rotatebox[origin=c]{270}{\hspace{-47mm} \small{Mustache} \hspace{11mm} \small{Eye Mask} }\\
    \end{minipage}
    \vspace{-3mm}
    \caption{We propose Stable Score Distillation (SSD), a method that improves text-guided editing by preserving original content structure and enhancing realism in edited results. The first column shows the original views, while the remaining columns display comparative results with existing score distillation methods on the SD model. SSD demonstrates superior performance, maintaining the integrity of the original content and producing more realistic edits.}
    \label{fig:teaser}
    \vspace{-3mm}
}

\maketitle

\begin{abstract}
Text-guided image and 3D editing have advanced with diffusion-based models, yet methods like Delta Denoising Score often struggle with stability, spatial control, and editing strength. These limitations stem from reliance on complex auxiliary structures, which introduce conflicting optimization signals and restrict precise, localized edits.
We introduce Stable Score Distillation (SSD), a streamlined framework that enhances stability and alignment in the editing process by anchoring a single classifier to the source prompt. Specifically, SSD utilizes \revision{Classifier-Free Guidance (CFG)} equation to achieves cross-prompt alignment, and introduces a constant term null-text branch to stabilize the optimization process.
This approach preserves the original content’s structure and ensures that editing trajectories are closely aligned with the source prompt, enabling smooth, prompt-specific modifications while maintaining coherence in surrounding regions. Additionally, SSD incorporates a prompt enhancement branch to boost editing strength, particularly for style transformations.
Our method achieves state-of-the-art results in 2D and 3D editing tasks, including NeRF and text-driven style edits, with faster convergence and reduced complexity, providing a robust and efficient solution for text-guided editing. Code is available: \url{https://github.com/Alex-Zhu1/SSD}.
\end{abstract}

\begin{figure*}
    \centering
    \includegraphics[width=0.99\textwidth]{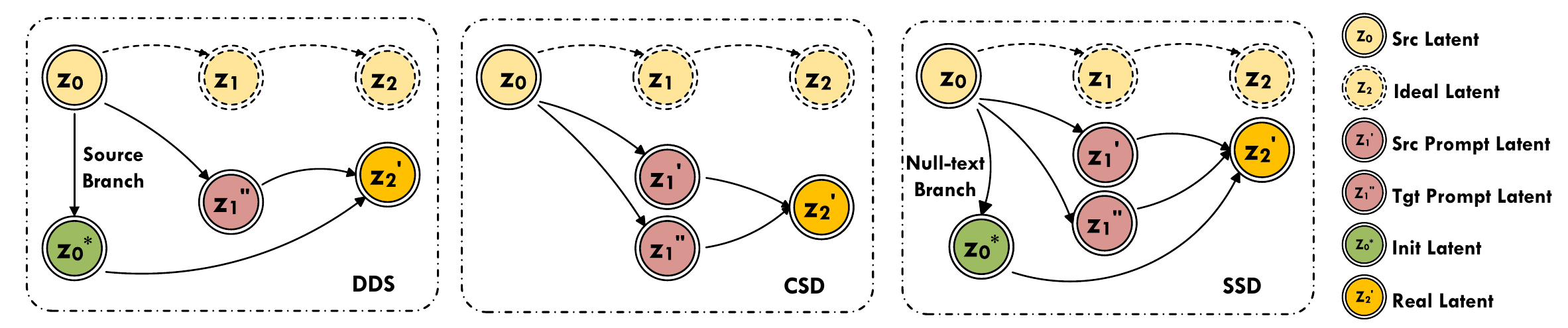}
    \vspace{-2.5mm}
    \caption{Illustration of three distillation-based approaches. \textit{Note that we assume a 2-step optimization process for illustration, where the subscript \( t \) represents the iteration number.} \rev{\textbf{DDS} utilizes the source branch to obtain initial latent \( Z_0^\ast \), while \textbf{CSD} employs two classifiers to derive \( Z_1' \) and \( Z_1'' \) for cross-prompt editing. Our \textbf{SSD} method designs a CFG classifier to determine the cross-prompt editing, introduces the null-text branch as the initial latent \( Z_0^\ast \), and further constructs the cross-trajectory term (see Sec.~\ref{subsec:ssd}) for stable optimization.}}
    \label{fig:diagram}
    \vspace{-5mm}
\end{figure*}

\begin{figure}[htbp]
    \centering
    \captionsetup[subfloat]{labelformat=empty,justification=centering}
    
    \begin{minipage}[c]{0.05\linewidth}
        \rotatebox{90}{\small DDS}
    \end{minipage}%
    \begin{minipage}[c]{0.97\linewidth}
        \includegraphics[width=0.24\linewidth]{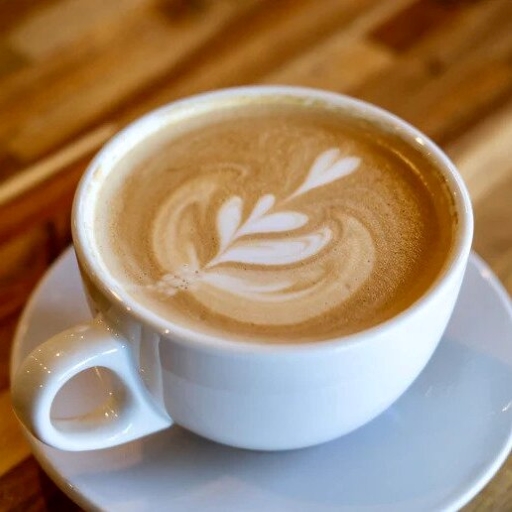}%
        \hspace{0.005\linewidth}
        \includegraphics[width=0.24\linewidth]{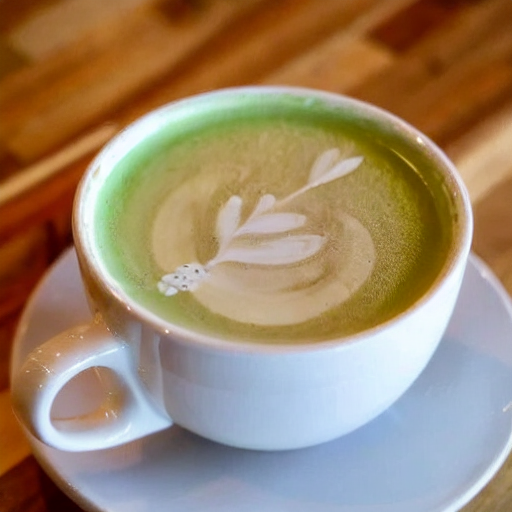}%
        \hspace{0.005\linewidth}
        \includegraphics[width=0.24\linewidth]{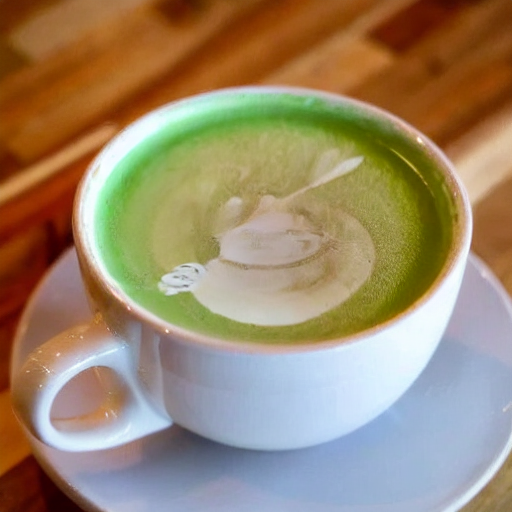}%
        \hspace{0.005\linewidth}
        \includegraphics[width=0.24\linewidth]{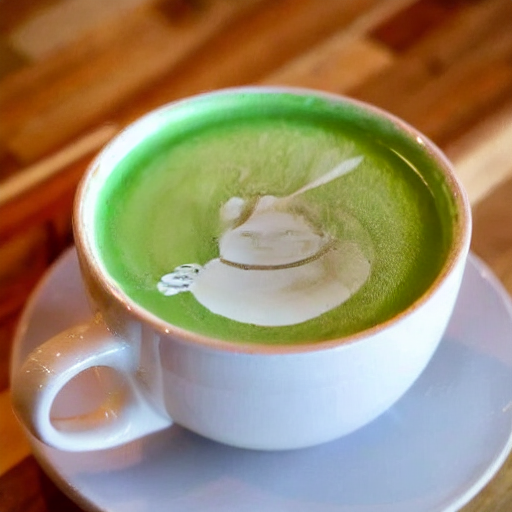}%
    \end{minipage}


    \begin{minipage}[c]{0.05\linewidth}
        \rotatebox{90}{\small SSD}
    \end{minipage}%
    \begin{minipage}[c]{0.97\linewidth}
        \includegraphics[width=0.24\linewidth]{figure/gradual/coffee.jpg}%
        \hspace{0.005\linewidth}
        \includegraphics[width=0.24\linewidth]{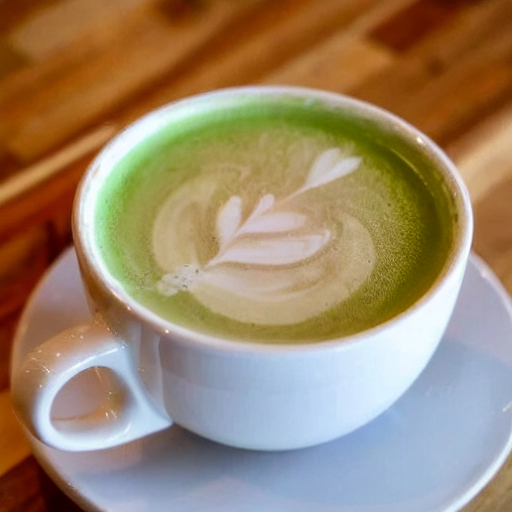}%
        \hspace{0.005\linewidth}
        \includegraphics[width=0.24\linewidth]{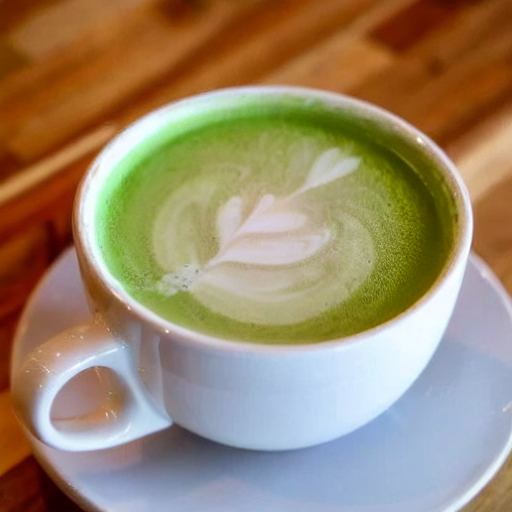}%
        \hspace{0.005\linewidth}
        \includegraphics[width=0.24\linewidth]{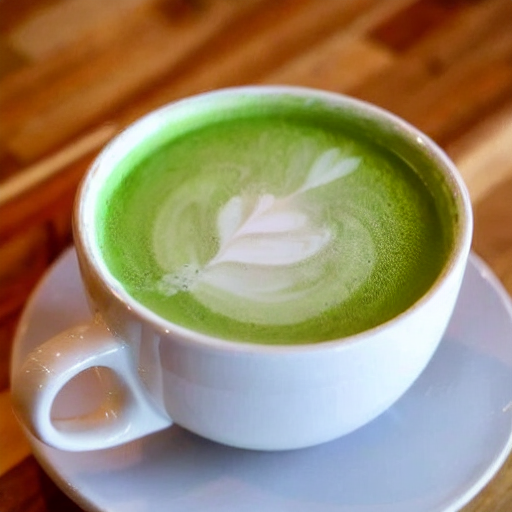}%
    \end{minipage}

    \vspace{1mm}

    \begin{minipage}{0.10\linewidth}
        \small 
        Iter:
    \end{minipage}%
    \begin{minipage}{0.9\linewidth}
        \small
        \hspace{10pt} 0 \hspace{50pt} 79 \hspace{40pt} 139 \hspace{40pt} 199
    \end{minipage}


    \begin{minipage}{0.97\linewidth}
        \centering
        \small \textnormal{\textit{A cup of \textcolor{blue}{coffee} $\rightarrow$ \textcolor{blue}{matcha}}}
    \end{minipage}%
    \vspace{2mm} 
    \hspace{0.02\linewidth}




    \vspace{-3mm}
    \caption{The optimization process of DDS and our SSD. SSD preserves the source structure effectively during optimization iterations, while DDS cannot preserve it effectively.}
    \vspace{-5mm}
    \label{fig:com_opti_dds}
\end{figure}

\section{Introduction}
\label{sec:intro}

Text-based image generation has achieved remarkable progress, particularly with the advent of diffusion models~\cite{ho2020denoising, song2021denoising, ramesh2022hierarchical, saharia2022photorealistic, rombach2022high}. These models leverage strong priors to produce high-quality images, facilitating significant advances in text-to-3D generation~\cite{poole2023dreamfusion, wang2024prolificdreamer, zhu2024hifa}. Moreover, text-guided 3D editing has enabled intricate modifications to shape~\cite{metzer2023latent} and texture~\cite{li2025learning}, supporting flexible and precise 3D scene manipulation.

Unlike generation tasks that create new content, editing tasks aim to modify specific elements within an image while preserving surrounding areas. However, directly applying methods like Score Distillation Sampling (SDS) to editing tasks can yield undesired effects, such as blurring across the image. This arises because SDS optimizes globally to the prompt, affecting regions beyond the targeted area~\cite{hertz2023delta, katzir2024noisefree}. DDS~\cite{hertz2023delta} addresses this by introducing a dual-branch architecture, pairing the source image with its description to leverage the model’s inherent bias and isolate specific prompt changes. Further, CSD~\cite{yu2024texttod} achieved scene editing by incorporating a classifier component within Classifier-Free Guidance (CFG)~\cite{ho2021classifierfree} to refine the prediction score by applying the classifier to both the source and target prompts. NFSD~\cite{katzir2024noisefree} further decomposes the CFG score, highlighting the classifier as the primary driver of prompt direction.

Despite their success, we argue that current distillation-based approaches face inherent limitations, such as low editing quality and loss of source content. \rev{As shown in Fig.~\ref{fig:diagram}, DDS~\cite{hertz2023delta} relies on the source branch to remove model bias, but lacks the explicit guidance to preserve the source content~\cite{koo2024posterior} during optimization.} As shown in Fig.~\ref{teaser:dds}, DDS changes man's clothe during editing his faces. 
Additionally, although introducing source prompt components is intended to improve prompt specificity~\cite{ju2024pnp}, it can amplify noise and introduce overlapping objectives that hinder stable convergence. This results in artifacts or unintended variations, especially in the unedited regions. Correspondingly, CSD~\cite{yu2024texttod} utilizes dual classifiers to refine the prompt editing direction (see Fig.~\ref{fig:diagram}). However, it lacks the explicit source preservation to restrict edits precisely to the target areas. As shown in Fig.~\ref{teaser:csd}, this causes the structure deformation and annoying artifacts in the edited regions.

Our insights into these limitations lead to two key observations: (1) \textbf{Cross-prompt}: a single classifier, providing the editing direction from source prompt to target, and (2) \textbf{Cross-trajectory}: stability in the editing process can be achieved by aligning the editing direction closely with the structure of the source content.

\rev{In this paper, we propose Stable Score Distillation (SSD), a streamlined approach for stable and precise text-guided editing. To achieve a smooth editing direction, we employ the CFG equation for both the source and target prompts, ensuring a gradual transition of the original contextual texture as the model adapts to the specified changes.
This approach contrasts with DDS~\cite{hertz2023delta}, as it eliminates the need for a auxiliary source branch, enabling our method to focus editing gradients precisely within target regions while ensuring a stable transition, as illustrated in Fig.~\ref{fig:com_opti_dds}. 
Moreover, for aligning the editing direction with the source prompt, facilitating smoother and more controlled progression toward the target prompt, we design an cross-trajectory strategy to ensure that edits respect the original structure, supporting subtle and stable transformations within designated areas.}
While NFSD~\cite{katzir2024noisefree} utilizes negative-branch and DDS utilizes source branch to enhance output clarity, as shown in SSD in Fig.~\ref{fig:diagram}, we introduce a null-text branch aligned with the ``no-edit'' direction to integrate a ``reconstruction'' term to explicitly enforce source content preservation, which enhances consistency and produces reliable edits across diverse tasks. Based on above designs, our framework remains streamlined and efficient, achieving both precision and stability without the complexity of additional components.

Our framework integrates seamlessly into existing DDS-based editing pipelines and applications, such as text-driven NeRF editing~\cite{koo2024posterior, park2024ednerf, kim2024dreamcatalyst} and 2D image editing~\cite{nam2024contrastive}. Notably, our approach’s ``clear'' editing direction preserves source content, making a carefully designed identity regularization~\cite{kim2024dreamcatalyst} unnecessary. Moreover, standard DDS methods often lack sufficient editing strength, resulting in minimal or negligible changes in output, particularly in style editing~\cite{kompanowski2024dream}. Our approach, with its streamlined and stable framework, allows for the seamless integration of a prompt enhancement branch to amplify editing capability.

With these improvements, our method achieves faster and more effective edits during optimization, remains compatible with the Stable Diffusion Model~\cite{rombach2022high} without requiring LoRA~\cite{hu2022lora} or fine-tuning, and integrates effectively with Instructpix2pix~\cite{brooks2023instructpix2pix}. Additionally, by incorporating non-increasing timestep sampling~\cite{huang2023dreamtime}, we accelerate convergence, reducing the required iterations to approximately 3,000 for NeRF~\cite{mildenhall2020nerf} and 1,500 for Gaussian splatting~\cite{kerbl20233d}.

In summary, our contributions are as follows:
\begin{itemize}
    \item We introduce a novel editing framework, Stable Score Distillation, that leverages a single, anchored classifier to achieve targeted and stable edits in 3D scene editing.
    \item We introduce a prompt enhancement strategy, effectively improve the prompt-alignment, especially style editing in 2D-image editing.
    \item We demonstrate the effectiveness of our approach across NeRF-editing and image-editing tasks, achieving state-of-the-art results with a streamlined and efficient framework.
\end{itemize}

\section{Related Work}

\subsection{Diffusion Models}

Diffusion models~\cite{ho2020denoising, song2021denoising,ren2025turbo2k,ren2024ultrapixel} have made significant advancements in generating diverse and high-fidelity images. Starting form a gaussian noise, diffusion models can predict the noise-less sample at each time step, until finally obtaining clear samples. Commonly, the denoising process can utilize U-net model to predict the noise. Some works~\cite{ho2020denoising, song2021scorebased} have observed that is proportional to the predicted score function~\cite{hyvarinen2005estimation} of the smoothed density. Thus, intuitively, taking steps in the direction of the score function gradually moves the sample towards the data distribution. 

To generate images aligned with a target prompt, guidance is typically introduced to explicitly control the weight assigned to the conditioning information. The popular guidance methods include Classifier Guidance~\cite{dhariwal2021diffusion} and Class-free Guidance (CFG)~\cite{ho2021classifierfree}. While the former rely on a separately learned classifier, the latter directly introduces null-text samples to the model. CFG modifies the score function to steer the process towards regions with a higher ratio of conditional density to the unconditional one. However, it has been observed that CFG trades sample fidelity for diversity~\cite{ho2021classifierfree}. 
Based on the insights gained from the decomposition of the CFG equation, we propose a novel Stable Score Distillation (SSD) method to \emph{guide} the SDS optimization process in Sec.~\ref{sec:method}.

\subsection{Score Distillation Sampling (SDS)}

Benefit from the data scale-law, diffusion model~\cite{rombach2022high, saharia2022photorealistic, ramesh2022hierarchical} achieve high-quality image generation and text-to-image generation. Specifically, Score Distillation Sampling (SDS)~\cite{poole2023dreamfusion} leveraging the priors of pre-trained text-to-image models to facilitate text-conditioned generation in 3D content generation. Specifically, SDS is an optimization approach that updates the rendering parameter towards the image distribution of diffusion models by enforcing the noise prediction on noisy rendered images to match sampled noise. While SDS provides an elegant mechanism for leveraging pretrained text-to-image models, SDS-generated results often suffer from oversaturation and lack of fine realistic details. VSD~\cite{wang2024prolificdreamer} proposed a particle-based optimization framework that  treats the 3D parameter as a random variable of target distribution. Furthermore, by regarding SDS as a reverse diffusion process, decreasing timesteps sampling~\cite{huang2023dreamtime, zhu2024hifa} to imitate the diffusion reverse sampling, which can improve the quality of the generated 3D assets.

In image editing, Delta Denoising Score (DDS)~\cite{hertz2023delta} found that Score Distillation Sampling (SDS) introduces noticeable artifacts and over-smoothing in edited images due to inherent bias. To mitigate this bias, DDS employs a subtraction of two SDS scores of the source and target images to obtain a delta score, which is then used to guide the optimization process.

\subsection{Text-Driven 3D-Scene Editing}

Text-driven 3D scene editing has been a popular research topic. IN2N~\cite{instructnerf2023} proposed a Iterative Dataset Update method that can edit 3D scenes from text descriptions. By leveraging advancements in 2D diffusion editing techniques, notably InstructP2P~\cite{brooks2023instructpix2pix} and ControlNet~\cite{zhang2023adding}, GaussianEditor\cite{chen2024gaussianeditor} and GaussCtrl~\cite{wu2024gaussctrl} utilize edit multi-view images latent to optimize the 3D scene. We consider utilize score distillation to guide the 3D scene editing, which is more flexible and efficient for the text-driven 3D scene editing.

Building upon the foundational SDS loss introduced by DreamFusion~\cite{poole2023dreamfusion}, some work has explore SDS loss in the text-driven 3D scene editing. RePaint-NeRF~\cite{RePaint-NeRF} has advanced the application of SDS in 3D editing by integrating a semantic mask to guide and constrain modifications within the background elements. CSD~\cite{yu2024texttod} utilize two classifiers to achieve editing. In a similar vein, ED-NeRF~\cite{park2024ednerf} has introduced an enhanced loss function specifically designed for 3D editing tasks. PDS~\cite{koo2024posterior} proposed a posterior distillation sampling to match stochastic latent~\cite{huberman2024edit}. {Piva~\cite{le2024preserving} fine-tuned the model while introducing a regularization term to preserve identity.} Unfortunately, these methods are still limited to the long-time diffusion reverse sampling process, which is not suboptimal for the text-driven 3D scene editing. DreamCatalyst~\cite{kim2024dreamcatalyst} extends the PDS optimization processing to ID-preserving and edit-ability based on decreasing timesteps sampling. 

Different from the above methods, ours firstly improve the DDS optimization process by introducing a single classifier, and further introduce a null-text branch to achieve a more stable and precise editing process.

\section{Preliminary}

\begin{figure*}
  \centering
  \centering
  \hspace{-1.5mm}
  \includegraphics[width=\textwidth]{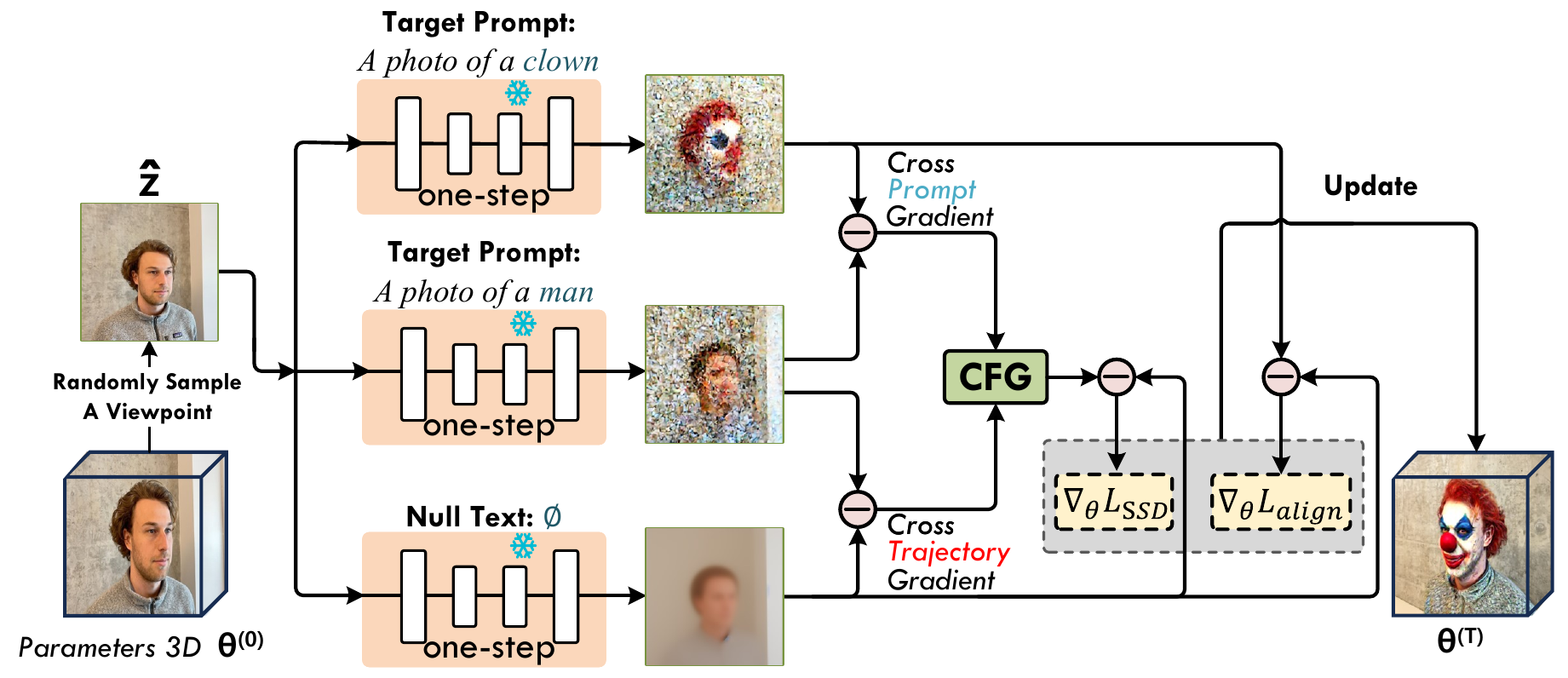}
  \vspace{-5mm}
  \caption{The overview of SSD. Given the parameter 3D-model or image, SSD provides effective editing gradient to guide the optimization process. We utilize CFG equation between the predicted target noise \( \epsilon_{\phi}(z_t, y, t) \) and source noise \( \epsilon_{\phi}(z_t, \hat{y}, t)\), which generate the gradual editing direction. Furthermore, we introduce a null-text branch \( \epsilon_{\phi}(\hat{z}_t, \varnothing, t) \) to regularize the optimization process and achieve stable optimization. We further analyzing and decompose ours design term into three parts: cross-prompt, cross-trajectory, and prompt-enhance.}
  \label{fig:overview_combined}
  \vspace{-5mm}
\end{figure*}

In this section, we first discuss existing optimization-based approaches to handle parametric images. Then, we will introduce our novel parametric image editing method in Section~\ref{sec:method}.

\subsection{Score Distillation Sampling}
\label{subsec:sds}

Score Distillation Sampling (SDS)~\cite{poole2023dreamfusion} is proposed to generate parametric images by leveraging the 2D prior of pretrained text-to-image diffusion models. Specifically, given a pretrained diffusion model \(\epsilon_{\phi}\), SDS optimizes a set of parameters \(\theta\) of a differentiable parametric image generator \(g\), using the gradient of the loss \(L_{\text{SDS}}\) with respect to \(\theta\):
\begin{equation}
\nabla_{\theta} L_{\text{SDS}} = w(t) \left( \epsilon_{\phi}(z_t(x); y, t) - \epsilon \right) \frac{\partial x}{\partial \theta},
\label{eq:sds}
\end{equation}
where \(x = g(\theta)\) is an image rendered by \(\theta\), \(z_t(x)\) is obtained by adding a Gaussian noise \(\epsilon\) to \(x\) corresponding to the \(t\)-th timestep of the diffusion process, and \(y\) is a condition to the diffusion model. 
As Noise-Free Score Distillation (NFSD)~\cite{katzir2024noisefree} has shown, the score \(\epsilon_{\phi}(z_t(x); y, t)\) provides the direction in which this noised version of \(x\) should be moved towards a denser region in the distribution of real images.

\subsection{Delta Distillation Sampling}
\label{subsec:dds}



Although SDS get excellent generation ability, for editing task, an undesired component from the pretrained model, \(\delta_{\text{bias}}\), interferes with the process and causes the image to become smooth and blurry in some parts~\cite{hertz2023delta}.
Based on the observations that a matched source prompt \( \hat{y} \) and source latent \(\hat{z}_t\) can estimate the noisy direction \(\delta_{\text{bias}}\), thus, the DDS method aims to remove the \(\delta_{\text{bias}}\) by introducing source branch, as shown in Eq.~\ref{eq:dds-2}:
\begin{equation}
    \nabla_{\theta} L_{\text{DDS}} = \left( \epsilon^c_{\phi}(z_t, y, t) - \epsilon^c_{\phi}(\hat{z}_t,\hat{y}, t) \right) \frac{\partial z}{\partial \theta},
    \label{eq:dds-2}
\end{equation}
where \(\epsilon^c_{\phi}(z_t, y, t)\) and \(\epsilon^c_{\phi}(\hat{z}_t, \hat{y}, t)\) are pretrained model predictions \(\epsilon\), with the superscript \(c\) indicating the CFG results. Thus, DDS pushes the optimized image into the direction of the target prompt without the interference of the noise component, namely, \(\nabla_{\theta} L_{\text{DDS}} \approx \delta_{\text{text}}\). Obviously, \(\nabla_{ \delta_{\text{text}}}\) is contingent on classifier part from \( \epsilon^c_{\phi}(z_t, y, t) \) as discussed in CSD~\cite{yu2024texttod} and NFSD~\cite{katzir2024noisefree}. Note that in the following manuscript, we decompose the CFG results without the superscript \(c\) and and omit the timestep \(t\) for simplicity.

Further exploring prompt editing direction, CSD~\cite{yu2024texttod} method proposed a dual-classifier to refine the editing score and achieve more precise editing, as shown in Eq.~\ref{eq:csd}:
\begin{equation}
    \begin{split}
      \nabla_{\theta} L_{\text{CSD}} = & ( \, w_a \left( \epsilon_{\phi}(z_t, y) - \epsilon_{\phi}(z_t, \varnothing) \right) \\
      & - w_b \left( \epsilon_{\phi}(z_t, \hat{y}) - \epsilon_{\phi}(z_t, \varnothing) \right) ) \frac{\partial z}{\partial \theta} ,
    \end{split}
    \label{eq:csd}
  \end{equation}
while the \(\epsilon_{\phi}(z_t, y, t)\) and \(\epsilon_{\phi}(z_t,\hat{y}, t)\) are \textbf{current} latent \( z_t \) predictions for the target prompt \( y \) and source prompt \( \hat{y} \), respectively. \( w_a \) and \( w_b\) are weights of classifiers. Simply put, CSD aims to refine the prompt editing direction by determining the difference between two classifiers, which can be regarded as a cross-prompt term.


\section{Method}
\label{sec:method}

In 3D scene editing process, which requires consideration of both the target prompt and the original source content, we consider two key aspects: (1) smooth editing direction towards the target prompt and (2) and editing results respect the original structure. Based on these, in this section, we introduce our novel editing framework Stable Score Distillation.

\subsection{Stable Score Distillation}
\label{subsec:ssd}

Firstly, we introduce the design of a cross-prompt editing direction. As discussed about CSD method in Sec.~\ref{subsec:dds}, the key role in cross-prompt editing is to provide a smooth transition from the source prompt to the target. As the CFG guidance~\cite{ho2021classifierfree} steers the process towards regions with a higher ratio of conditional density to the unconditional one, accordingly, we can modify the SDS score function, as shown below:
\begin{equation}
    Grad = \epsilon_{\phi}(z_t, \hat{y}) + s \left( \epsilon_{\phi}(z_t, y) - \epsilon_{\phi}(z_t, \hat{y}) \right),
    \label{eq:ours_cfg}
\end{equation}
where \(\epsilon_{\phi}(z_t, y)\) and \(\epsilon_{\phi}(z_t, \hat{y})\) are pretrained model predictions. The scale factor \( s \) is equal to control weight.

Although the cross-prompt term provides a smooth texture transition in the edited region, we observed that the optimization process leads to abrupt structural changes, often resulting in artifacts and unappealing outcomes, similar to CSD in Fig.~\ref{teaser:csd}. To address this, we introduce an additional regularization term to constrain the structural transition.
Interestingly, as shown in Fig.~\ref{teaser:dds}, DDS achieves better results than CSD by incorporating a source branch. 
However, DDS still lacks a mechanism to ensure the original structure remains intact, leading to modification on unedited regions.
To address this, we introduce a null-text branch \( \epsilon_{\phi}(\hat{z}_t, \varnothing) \) to regularize the optimization process, as shown in Eq.~\ref{eq:our_full}:
\begin{equation}
        L_\text{ssd} =  \epsilon_{\phi}(z_t, \hat{y}) + s ( \epsilon_{\phi}(z_t, y) - \epsilon_{\phi}(z_t, \hat{y}) )  -  \epsilon_{\phi}(\hat{z}_t, \varnothing).
    \label{eq:our_full}
\end{equation}

Eq.~\ref{eq:our_full} is ours Stable Score Distillation, and we can further decompose above equation into two parts, and the latter is regarded as a cross-trajectory term. 
\begin{equation}
        L_\text{ssd} = \underbrace{w_p \left( \epsilon_{\phi}(z_t, y) - \epsilon_{\phi}(z_t, \hat{y}) \right)}_{\text{cross-prompt}}
        + \underbrace{ w_t \left( \epsilon_{\phi}(z_t, \hat{y}) - \epsilon_{\phi}(\hat{z}_t, \varnothing) \right)}_{\text{cross-trajectory}},
    \label{eq:our_ssd}
\end{equation}
where the \( w_t \) and \( w_p \) control the strength of the cross-trajectory and cross-prompt, respectively.

The cross-trajectory term can be interpreted as the distance between the transitions of two latents, ensuring that the original structure remains smooth and does not change abruptly (more details are provided in the supplementary material).
In Fig.~\ref{fig:scaling_terms}, we can see that the cross-trajectory term can provide a strong structure constraint ability, guiding the optimization process to preserve the source image structure. Specifically, when set \( w_t = 0 \), the optimization process behaves similarly to the CSD\cite{yu2024texttod} method, which fails to retain the original image structure.

\subsection{Improving Prompt Alignment}

\begin{figure}[t]
    \centering
    \setlength{\tabcolsep}{1pt} 
    \captionsetup[subfloat]{labelformat=empty,justification=centering}
    \hspace{-1.5mm}
    \subfloat[$w_t=0$]{
     \begin{minipage}{0.225\linewidth}
     \includegraphics[width=\linewidth]{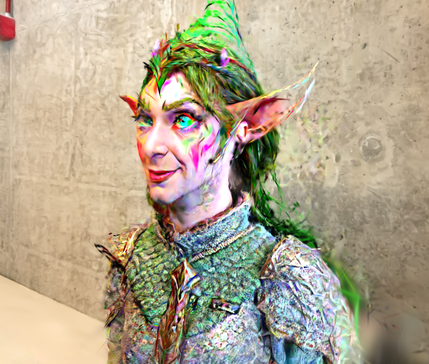}\vspace{0.6mm}
     \includegraphics[width=\linewidth]{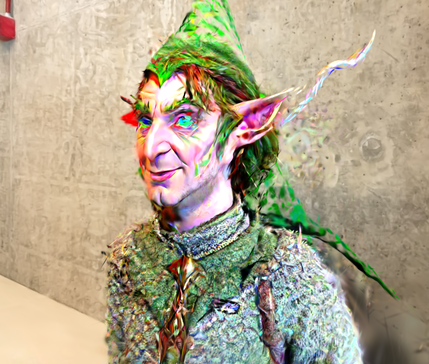}\vspace{0.6mm}
     \includegraphics[width=\linewidth]{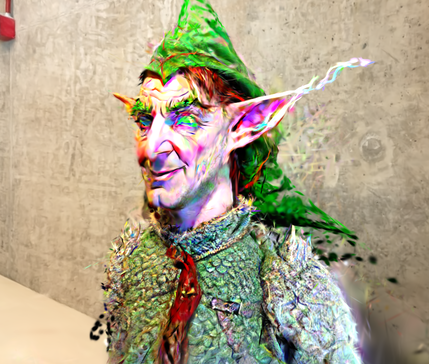}\vspace{0.6mm}
     \includegraphics[width=\linewidth]{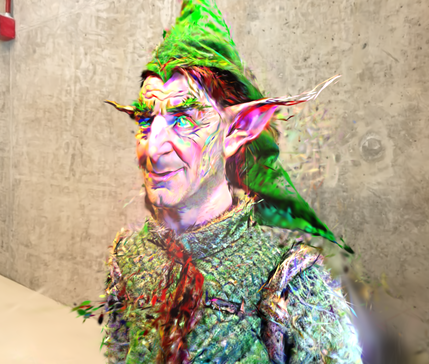}
     \end{minipage}
     }
    \hspace{-1.5mm}
    \subfloat[$w_t=1.0$]{
     \begin{minipage}{0.225\linewidth}
     \includegraphics[width=\linewidth]{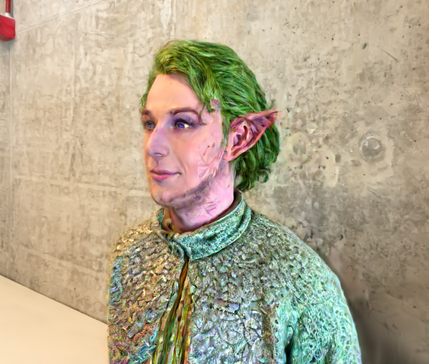}\vspace{0.6mm}
     \includegraphics[width=\linewidth]{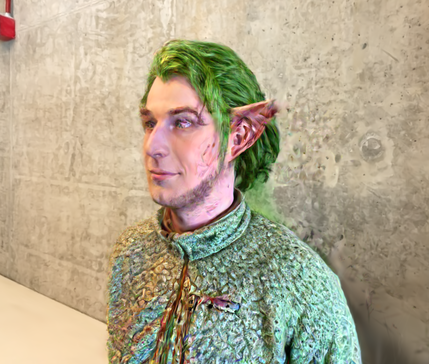}\vspace{0.6mm}
     \includegraphics[width=\linewidth]{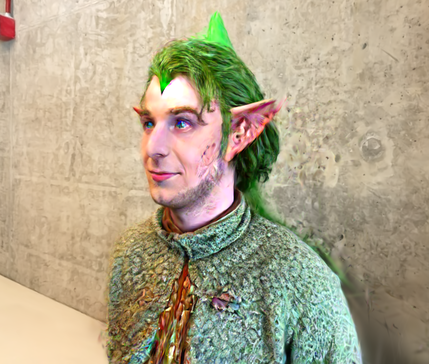}\vspace{0.6mm}
     \includegraphics[width=\linewidth]{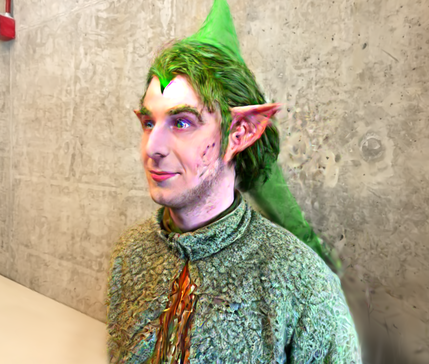}
     \end{minipage}
     }
    \hspace{-1.5mm}
    \subfloat[$w_t=1.5$]{
      \begin{minipage}{0.225\linewidth}
      \includegraphics[width=\linewidth]{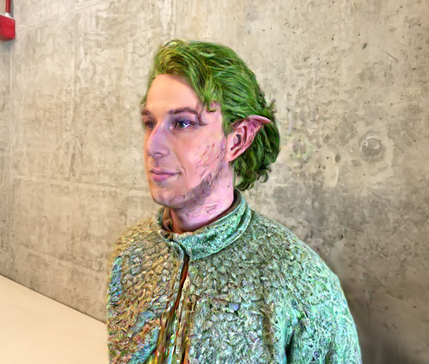}\vspace{0.6mm}
      \includegraphics[width=\linewidth]{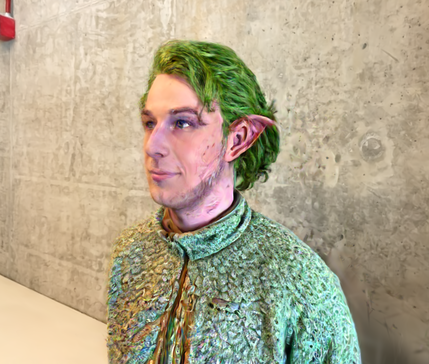}\vspace{0.6mm}
      \includegraphics[width=\linewidth]{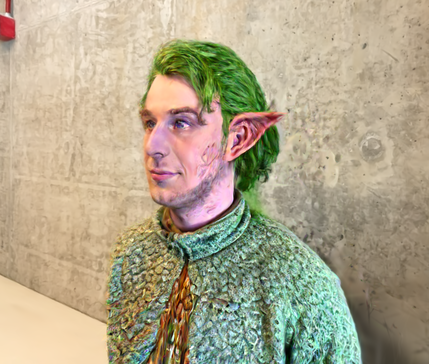}\vspace{0.6mm}
      \includegraphics[width=\linewidth]{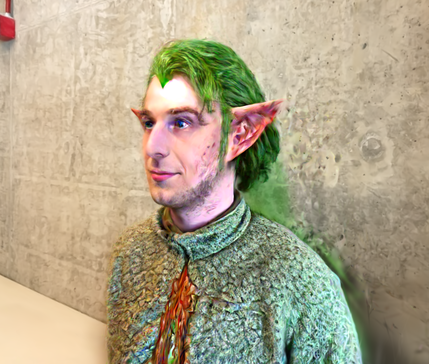}
      \end{minipage}
      }
    \hspace{-1.5mm}
    \subfloat[$w_t=2.0$]{
      \begin{minipage}{0.225\linewidth}
      \includegraphics[width=\linewidth]{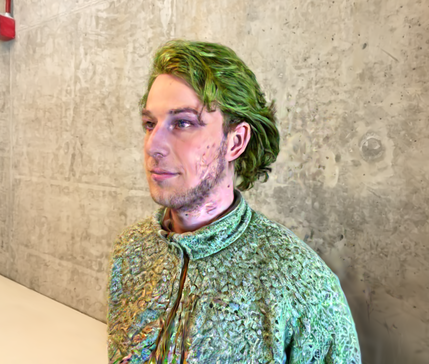}\vspace{0.6mm}
      \includegraphics[width=\linewidth]{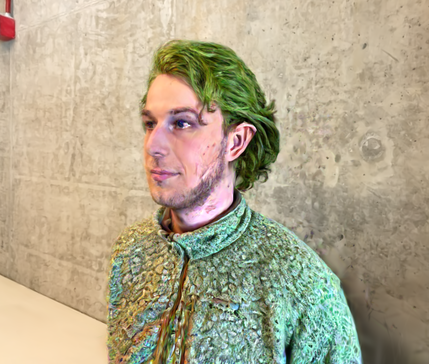}\vspace{0.6mm}
      \includegraphics[width=\linewidth]{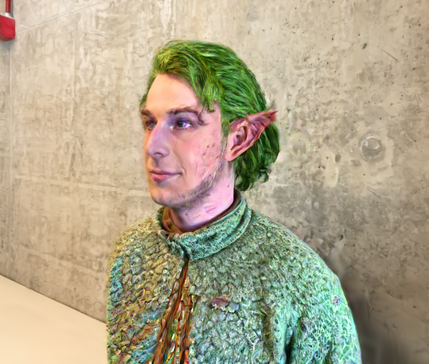}\vspace{0.6mm}
      \includegraphics[width=\linewidth]{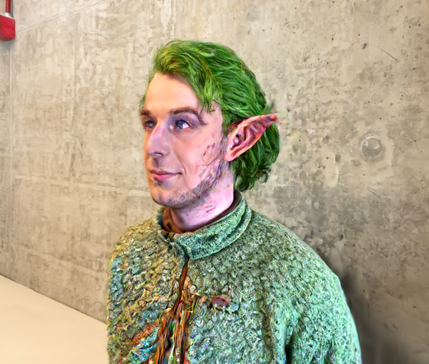}
      \end{minipage}
      }
    \hspace{-3.5mm}
    \begin{minipage}{0.08\linewidth}
        \vspace{-62mm}
        \hspace*{4mm}\rotatebox[origin=c]{270}{\scriptsize $w_e=0$}\\[5.5mm]
        \hspace*{4mm}\rotatebox[origin=c]{270}{\scriptsize $w_e=1.5$}\\[5.5mm]
        \hspace*{4mm}\rotatebox[origin=c]{270}{\scriptsize $w_e=5.5$}\\[5.5mm]
        \hspace*{4mm}\rotatebox[origin=c]{270}{\scriptsize $w_e=7.5$}
    \end{minipage}
    \caption{The effect of increasing the strength of the prompt enhancement term ($w_e$) and cross-trajectory term ($w_t$), with the cross-prompt term fixed at 7.5.
    Both terms contribute to prompt-aligned results, while setting ($w_t=0$) leads to saturation and discard source content.}
    \vspace{-3mm}
    \label{fig:scaling_terms}
\end{figure}

Although Eq.~\ref{eq:our_full} can achieve gradual editing results, we found Eq.~\ref{eq:our_full} have similar limitation with DDS~\cite{deutch2024turboedit}, which have insufficient editing strength. 
The editing results neither get successful editing nor retain the source image structure, often leads to little or no change in the final. Benefit from the cross-prompt editing design as Eq.~\ref{eq:our_full}, we can add a target prompt enhancement branch to guide the optimization process. The target prompt alignment branch will provide the direction of the target prompt, as shown in Eq.~\ref{eq:style_branch}:
\begin{equation}
        L_\text{align} = {w_e} \left( \epsilon_{\phi}(z_t, y) - \epsilon_{\phi}(z_t, \varnothing) \right) ,
    \label{eq:style_branch}
\end{equation}
where \( w_e \) is the prompt enhancement scale. As shown in Fig.~\ref{fig:scaling_terms}, the synchronous scaling of both the cross-trajectory and prompt-enhancement terms results in effective visual editing outcomes.

\subsection{Source Latent Regularization}

Empirically, we found that directly using latent-space loss rather than pixel-level loss can lead to optimization difficulties in local regions of 3DGS. For example, the bright spots appearing in Fig.~\ref{fig:scaling_terms}. To suppress the steep gradients in these areas, we incorporate ID regularization to guide the stable optimization process. Differ with PDS~\cite{koo2024posterior} use source latent \( \hat{x}_0 \), we can use the noisy latent \( \hat{x}_t \) to avoid partial exploding gradient, as shown in Eq.~\ref{eq:cross_traj}:
\begin{equation}
    L_\text{ID} = w(t) \cdot ( x_t - \hat{x}_t) ,
    \label{eq:cross_traj}
\end{equation}
where the \( w(t) \)  is the iteration-dependent strength, designed as a decreasing function of \( t \). Notably, the \( w(t) \) is not necessary to well-designed in our design.

Our final loss function as shown in Eq.~\ref{eq:final}:
\begin{equation}
    L_\text{final} =  L_\text{ssd} + L_\text{align} + L_\text{ID} .
    \label{eq:final}
\end{equation}

Based on the above design, we achieve a more prompt-aligned editing method, which integrates seamlessly into the Stable Diffusion Model~\cite{rombach2022high} without requiring LoRA~\cite{hu2022lora} or fine-tuning. Moreover, we will further introduce our method's connection with InstructPix2Pix~\cite{brooks2023instructpix2pix}.

\subsection{Connecting with IP2P}

The final design of our method is shown in Eq~\ref{eq:final}. We found that ours edit gard provide new angle to understand about InstructP2P~\cite{brooks2023instructpix2pix} one-step reverse sampling.
\begin{flalign}
    \begin{split}
        \epsilon_{\theta}(z_t, c_I, c_T) 
        &= \epsilon_{\theta}(z_t, \varnothing, \varnothing) \\
        &\quad + s_I \big(\epsilon_{\theta}(z_t, c_I, \varnothing) - \epsilon_{\theta}(z_t, \varnothing, \varnothing) \big) \\
        &\quad + s_T \big(\epsilon_{\theta}(z_t, c_I, c_T) - \epsilon_{\theta}(z_t, c_I, \varnothing) \big) ,
    \end{split} &&
    \label{eq:ip2p}
\end{flalign}
where \( c_I \) and \( c_T \) are input-image and instruction prompt separately,\( s_I \) and \( s_T \) are the source image control and instruction prompt control strength. The Eq.~\ref{eq:ip2p} is the InstructP2P one-step reverse sampling, which can provide the direction of the target prompt. We can see that the InstructP2P is the simple version of our method, the middle term of Eq.~\ref{eq:ip2p} is cross-trajectory regularization, and the last term of Eq.~\ref{eq:ip2p} is ours cross-prompt term. 
Simply put, as analyzing the Eq.~\ref{eq:our_full}, subtracting the constant correction term \( \epsilon_{\theta}(\hat{z}_t; \varnothing; \varnothing) \) is edit grad. Ours method reveal that apply DDS loss in the InstructP2P model can only editing branch, and don't have to provide the source branch. 
    
\section{Experiments}

In this section, we conduct editing experiments across two types of parameterized images. Section~\ref{subsec:exp-nerf} evaluates the effectiveness of our method on 3D Scenes Editing, and Section~\ref{subsec:exp-img} evaluates the effectiveness of our method on 2D Image Editing. We also conduct ablation studies to analyze the effectiveness of ours components in Section~\ref{subsec:exp-ab}.

\subsection{3D Scenes Editing}
\label{subsec:exp-nerf}

\textbf{Dataset.} To evaluate the effectiveness of our method, we conduct experiments on the scenes from IN2N~\cite{instructnerf2023} and other real-world datasets, including LLFF~\cite{mildenhall2019local} and Mip-Nerf360~\cite{barron2022mip}. 

\noindent \textbf{Baselines.} We compare our method with several state-of-the-art inversion methods. We use 3DGS~\cite{kerbl20233d} as the 3D representation, and compare our method with InstructNerf2Nerf~\cite{instructnerf2023}, DDS~\cite{hertz2023delta}, GS-Edit~\cite{chen2024gaussianeditor}, and DGE~\cite{chen2024dge}. For fairness, we implement the DDS version based on the official GS-Edit code. PDS~\cite{koo2024posterior} is designed for addition of objects to unspecified regions, we will provide the comparison results in supplementary material.

\noindent \textbf{Evaluation Metrics.} We follow common practice~\cite{instructnerf2023,chen2024gaussianeditor, chen2024dge} to evaluate the effectiveness of our method. CLIP Similarity is to evaluate the alignment between the render images and the target prompts, i.e.,  the cosine similarity between the text and image embeddings encoded by CLIP. Specifically, follow DGE~\cite{chen2024dge}, randomly sample 20 camera poses to evaluate. CLIP Directional Similarity is to measure the editing effect, i.e., the cosine similarity between the image and text editing directions (target embeddings minus source embeddings). 
We evaluate all methods on 6 different scenes and 10 different prompts.

\begin{figure*}[t]
\centering
    \captionsetup[subfloat]{labelformat=empty,justification=centering}

    \begin{minipage}[c]{0.95\linewidth}
        \centering
        \small
        \text{\textit{A photo of a \textcolor{blue}{man} $\rightarrow$ \textcolor{blue}{Spider man with a mask}}}
    \end{minipage}%

    \vspace{1pt}
    \begin{minipage}[c]{0.99\linewidth}

        \begin{minipage}{0.19\textwidth}
            \centering
            \begin{tikzpicture}
                \node at (0, 0) {\includegraphics[width=0.74\linewidth]{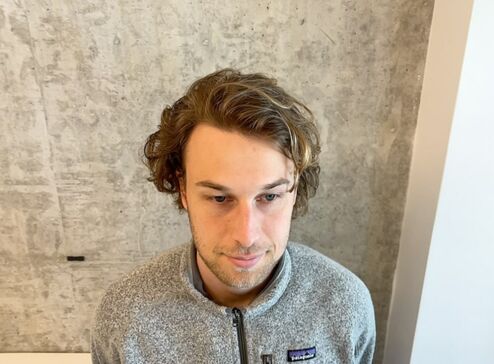}};
                \node at (1.1, -1.1) {\includegraphics[width=0.45\linewidth]{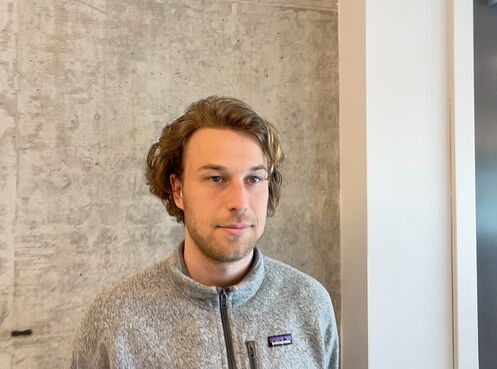}};
            \end{tikzpicture}
        \end{minipage}
        \begin{minipage}{0.19\textwidth}
            \centering
            \begin{tikzpicture}
                \node at (0, 0) {\includegraphics[width=0.74\linewidth]{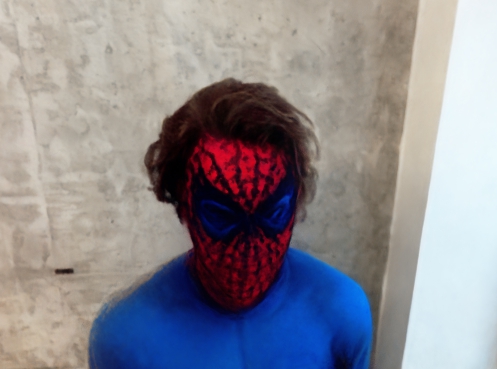}};
                \node at (1.1, -1.1) {\includegraphics[width=0.45\linewidth]{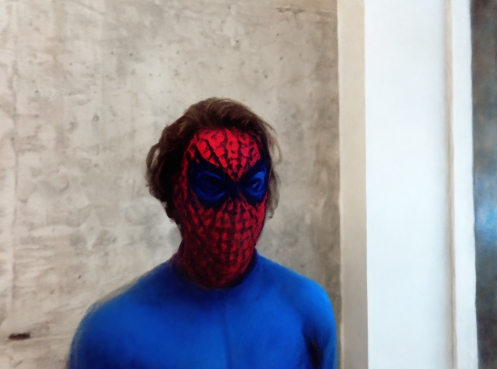}};
            \end{tikzpicture}
        \end{minipage}
        \begin{minipage}{0.19\textwidth}
            \centering
            \begin{tikzpicture}
                \node at (0, 0) {\includegraphics[width=0.74\linewidth]{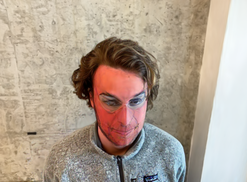}};
                \node at (1.1, -1.1) {\includegraphics[width=0.45\linewidth]{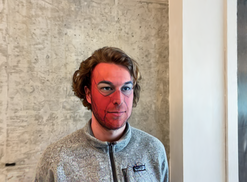}};
            \end{tikzpicture}
        \end{minipage}
        \begin{minipage}{0.19\textwidth}
            \centering
            \begin{tikzpicture}
                \node at (0, 0) {\includegraphics[width=0.74\linewidth]{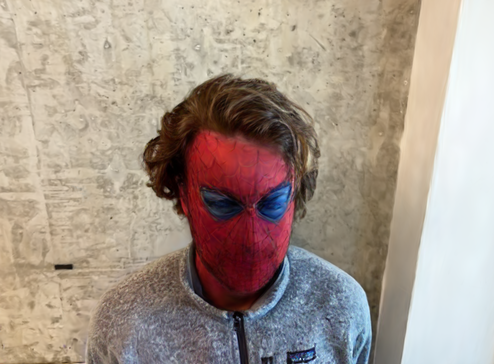}};
                \node at (1.1, -1.1) {\includegraphics[width=0.45\linewidth]{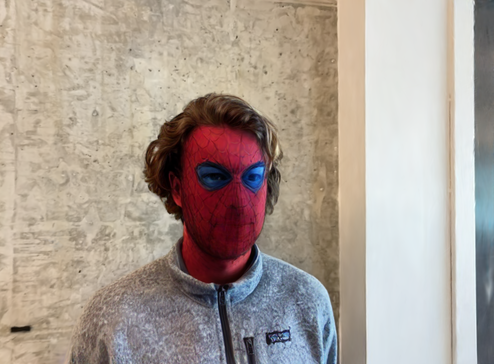}};
            \end{tikzpicture}
        \end{minipage}
        \begin{minipage}{0.19\textwidth}
            \centering
            \begin{tikzpicture}
                \node at (0, 0) {\includegraphics[width=0.74\linewidth]{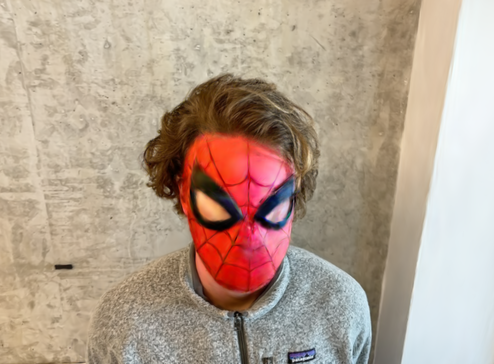}};
                \node at (1.1, -1.1) {\includegraphics[width=0.45\linewidth]{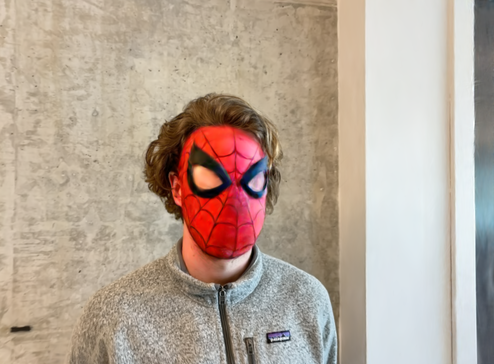}};
            \end{tikzpicture}
        \end{minipage}
    \end{minipage}

    \vspace{5pt}
    \hspace{0.01\linewidth}
    \begin{minipage}[c]{0.95\linewidth}
        \centering
        \small
        \text{\textit{A man wearing T-shirt \textcolor{blue}{with a pineapple pattern}}}
    \end{minipage}%

    \vspace{1pt}
    \begin{minipage}[c]{0.99\linewidth}
        \includegraphics[width=0.095\linewidth]{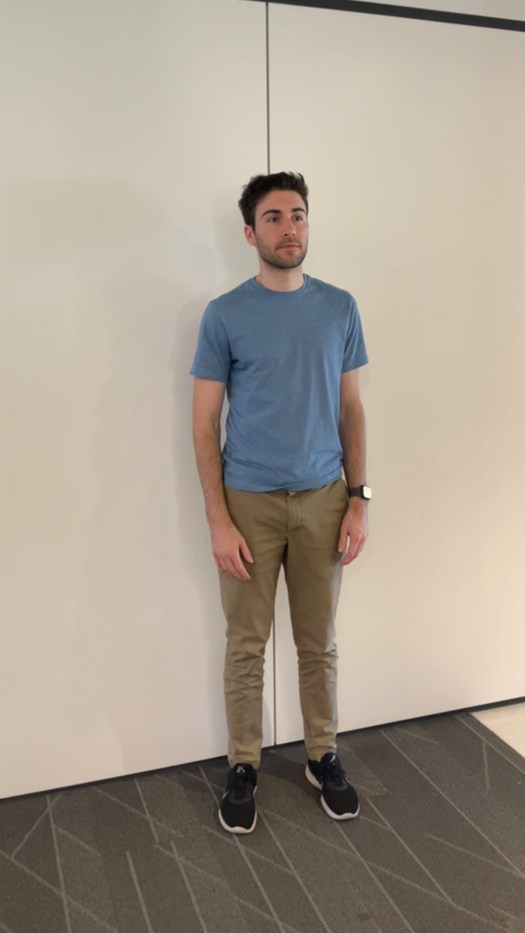}%
        \includegraphics[width=0.095\linewidth]{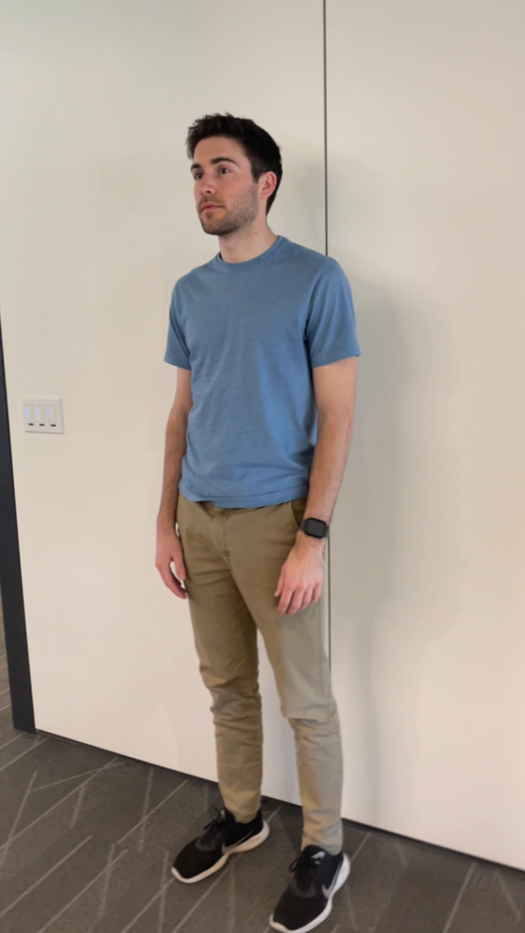}%
        \hspace{0.01mm}
        \includegraphics[width=0.095\linewidth]{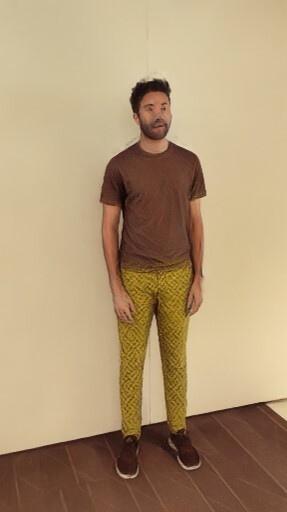}%
        \includegraphics[width=0.095\linewidth]{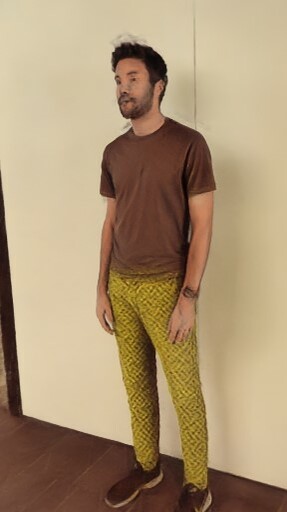}%
        \hspace{0.01mm}
        \includegraphics[width=0.095\linewidth]{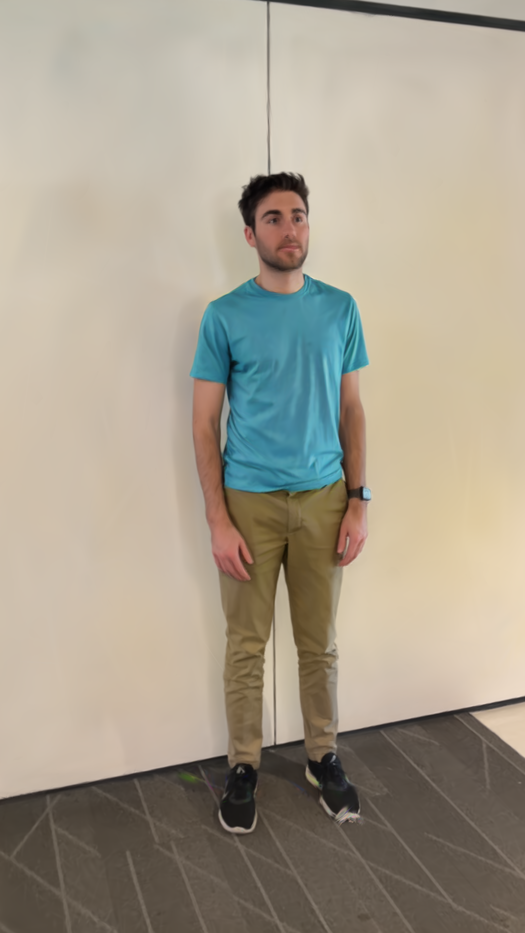}%
        \includegraphics[width=0.095\linewidth]{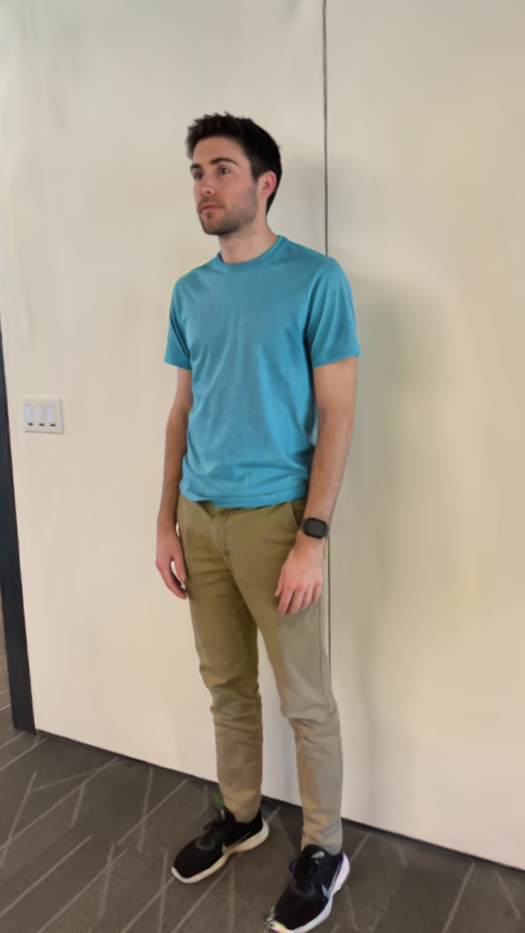}%
        \hspace{0.01mm}
        \includegraphics[width=0.095\linewidth]{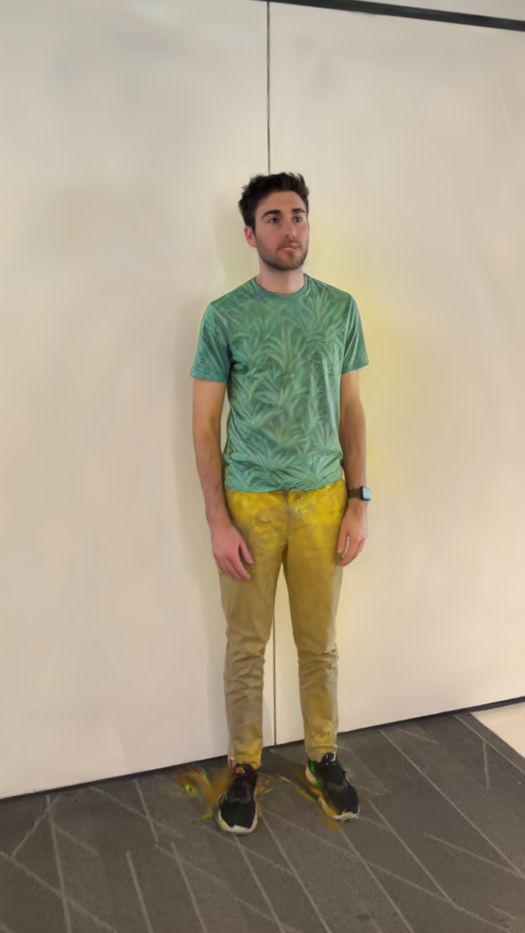}%
        \includegraphics[width=0.095\linewidth]{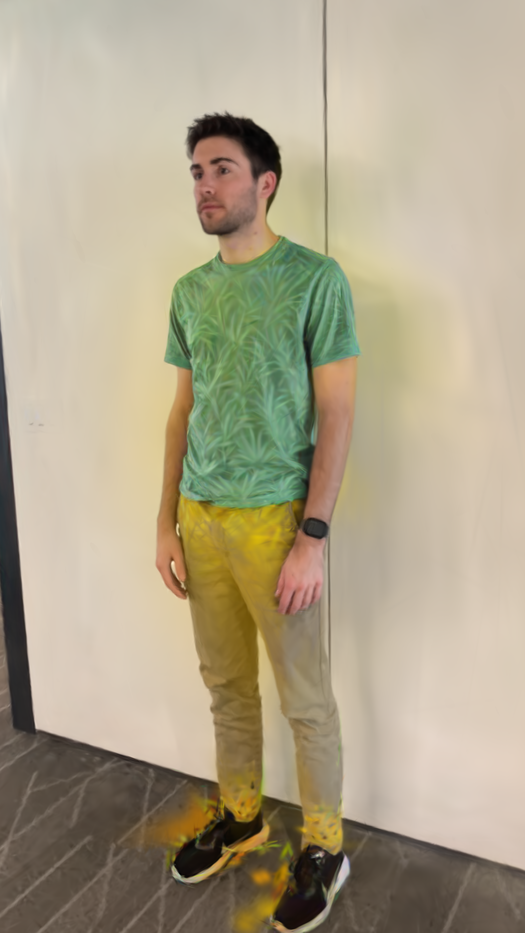}%
        \hspace{0.01mm}
        \includegraphics[width=0.095\linewidth]{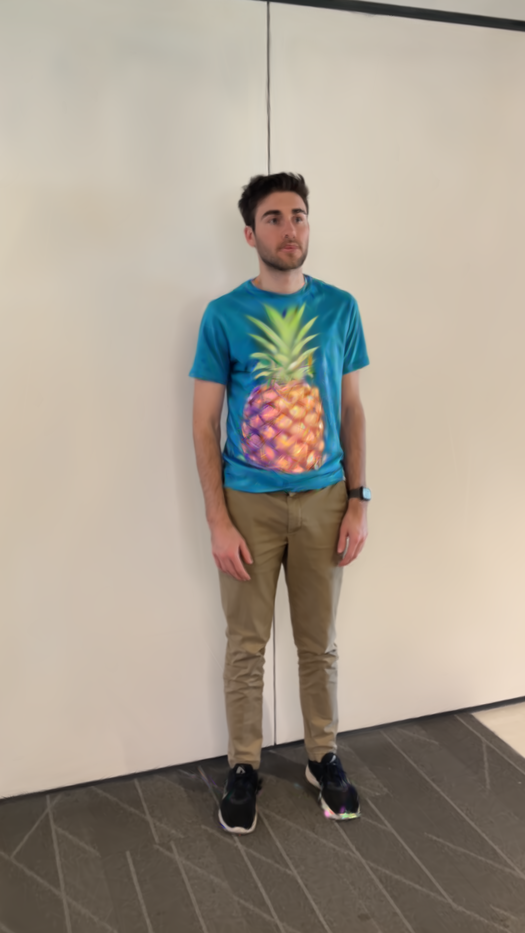}%
        \includegraphics[width=0.095\linewidth]{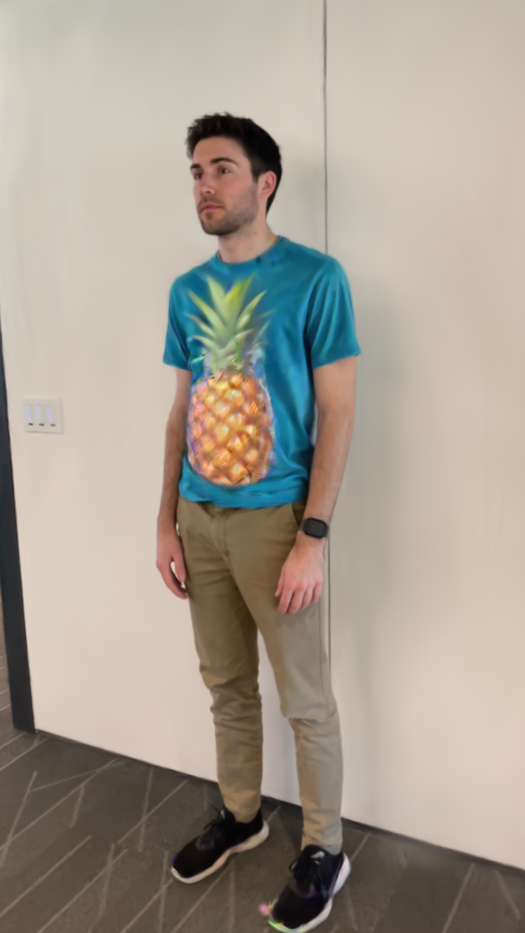}%
    \end{minipage}

    \vspace{5pt}
    \begin{minipage}[c]{0.95\linewidth}
        \centering
        \small
        \text{\textit{\textcolor{blue}{Rainbow} horn fossil}}
    \end{minipage}%

    \vspace{1pt}
    \begin{minipage}[c]{0.99\linewidth}
        \includegraphics[width=0.19\linewidth]{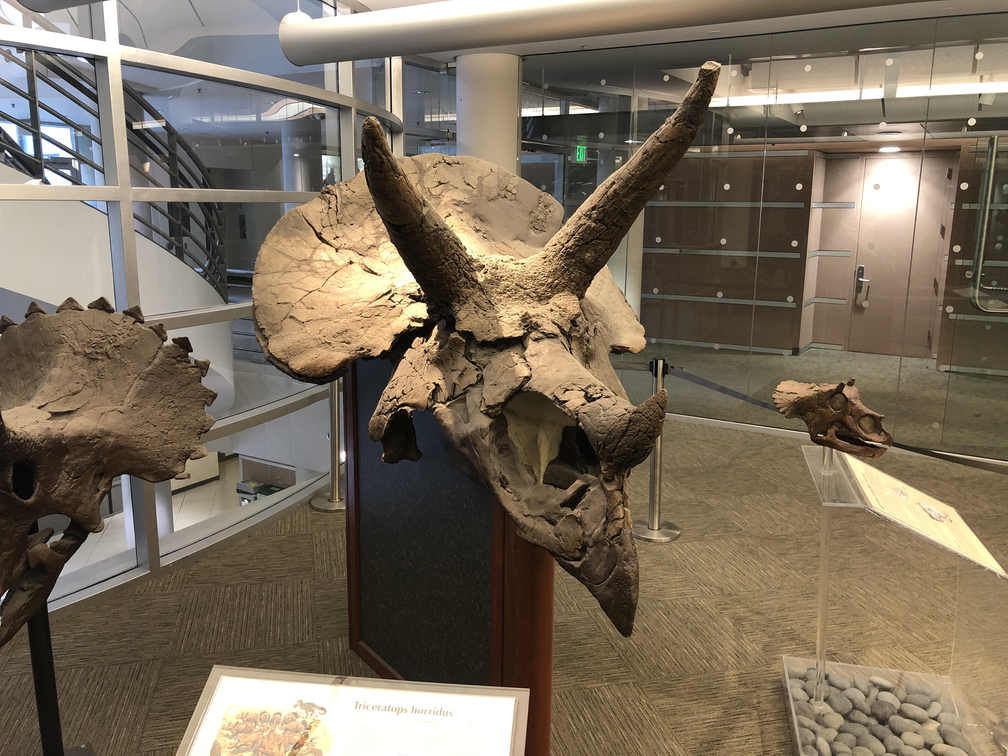}%
        \hspace{0.01mm}
        \includegraphics[width=0.19\linewidth]{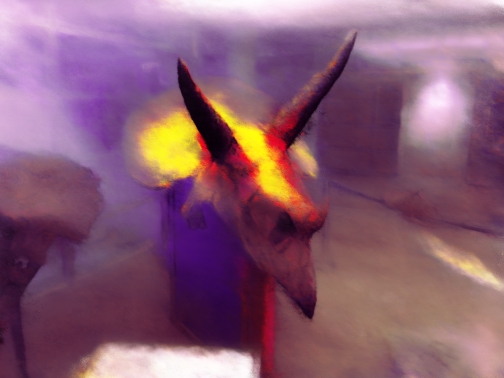}%
        \hspace{0.01mm}
        \includegraphics[width=0.19\linewidth]{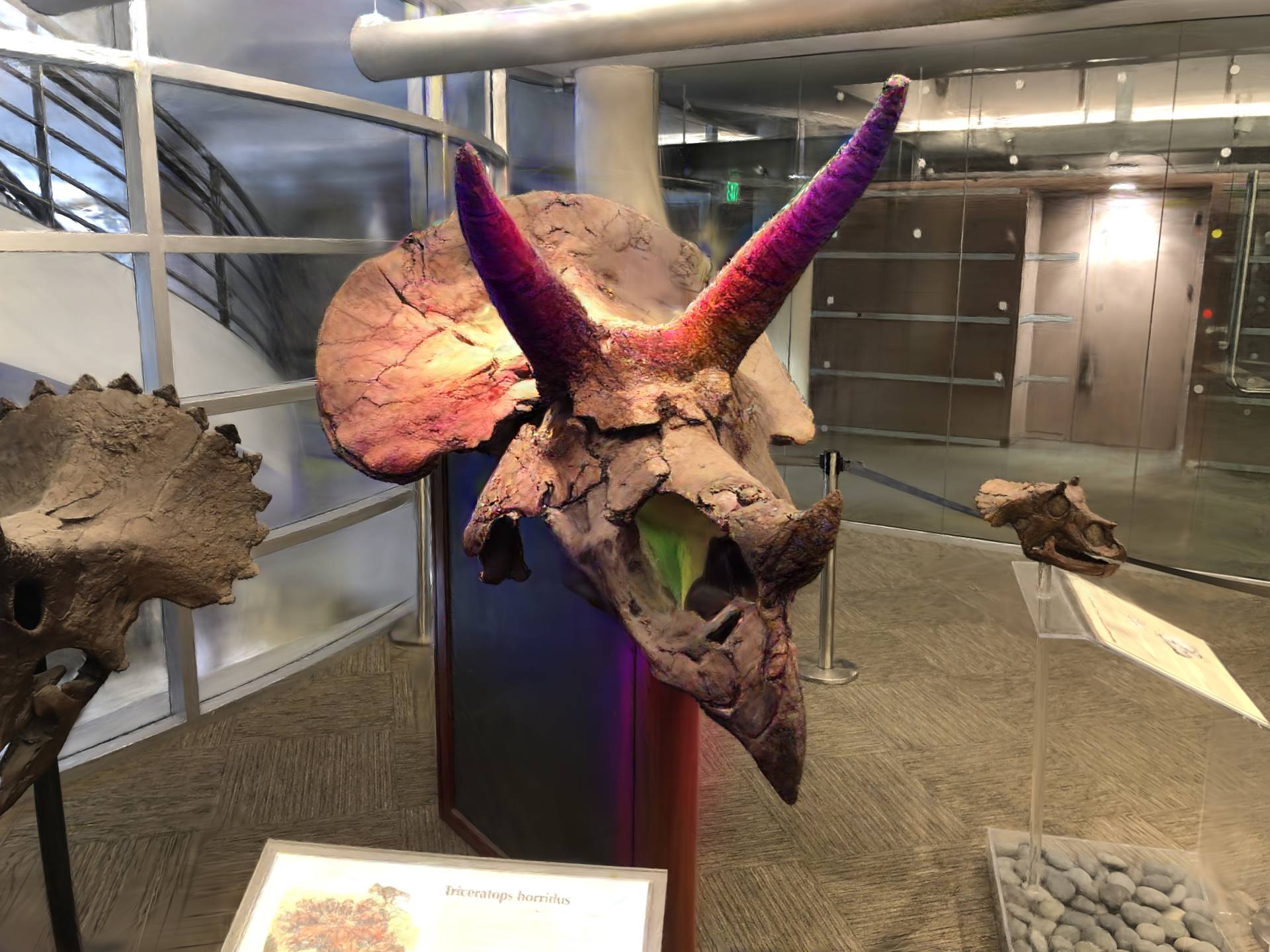}%
        \hspace{0.01mm}
        \includegraphics[width=0.19\linewidth]{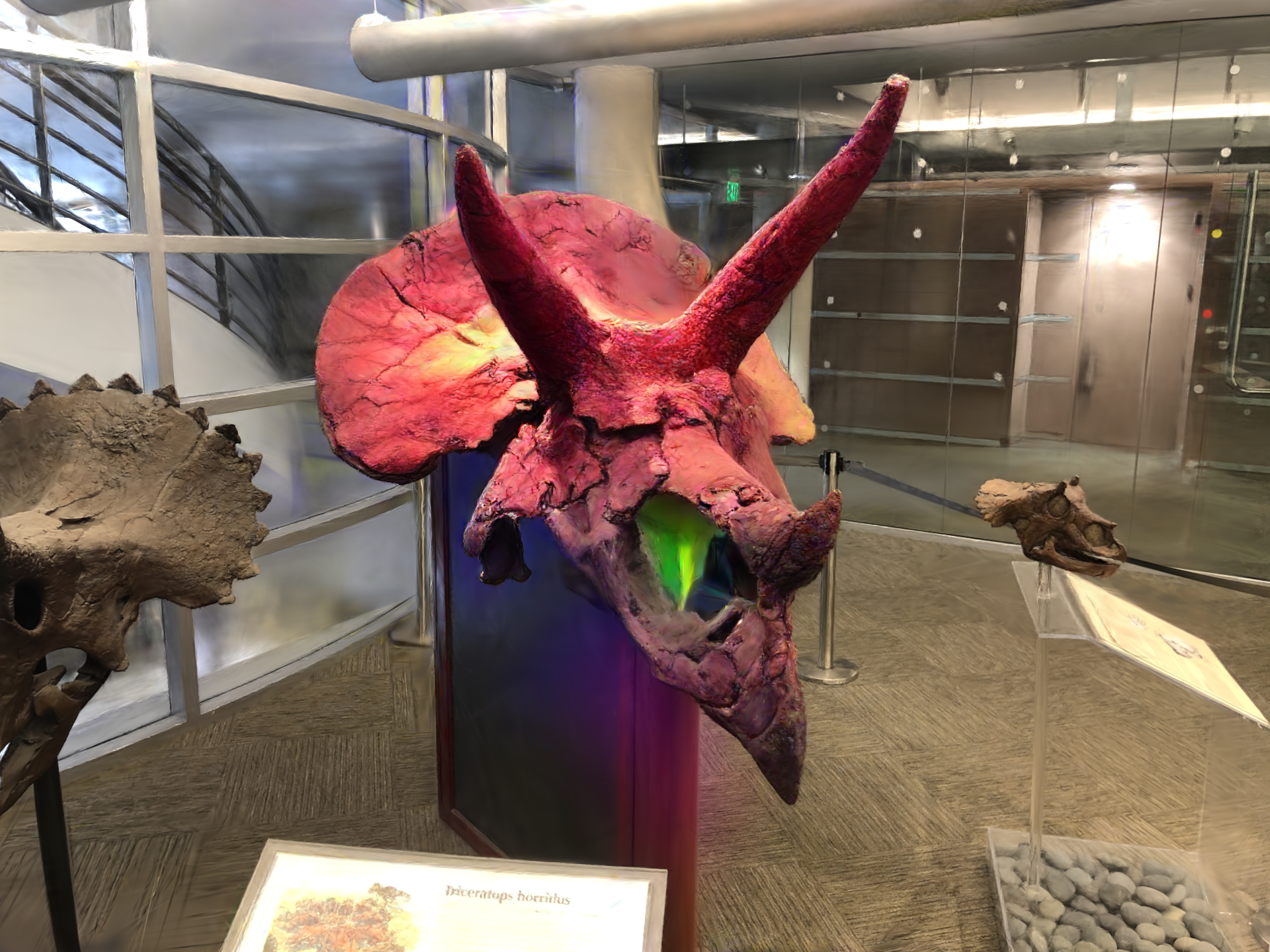}%
        \hspace{0.01mm}
        \includegraphics[width=0.19\linewidth]{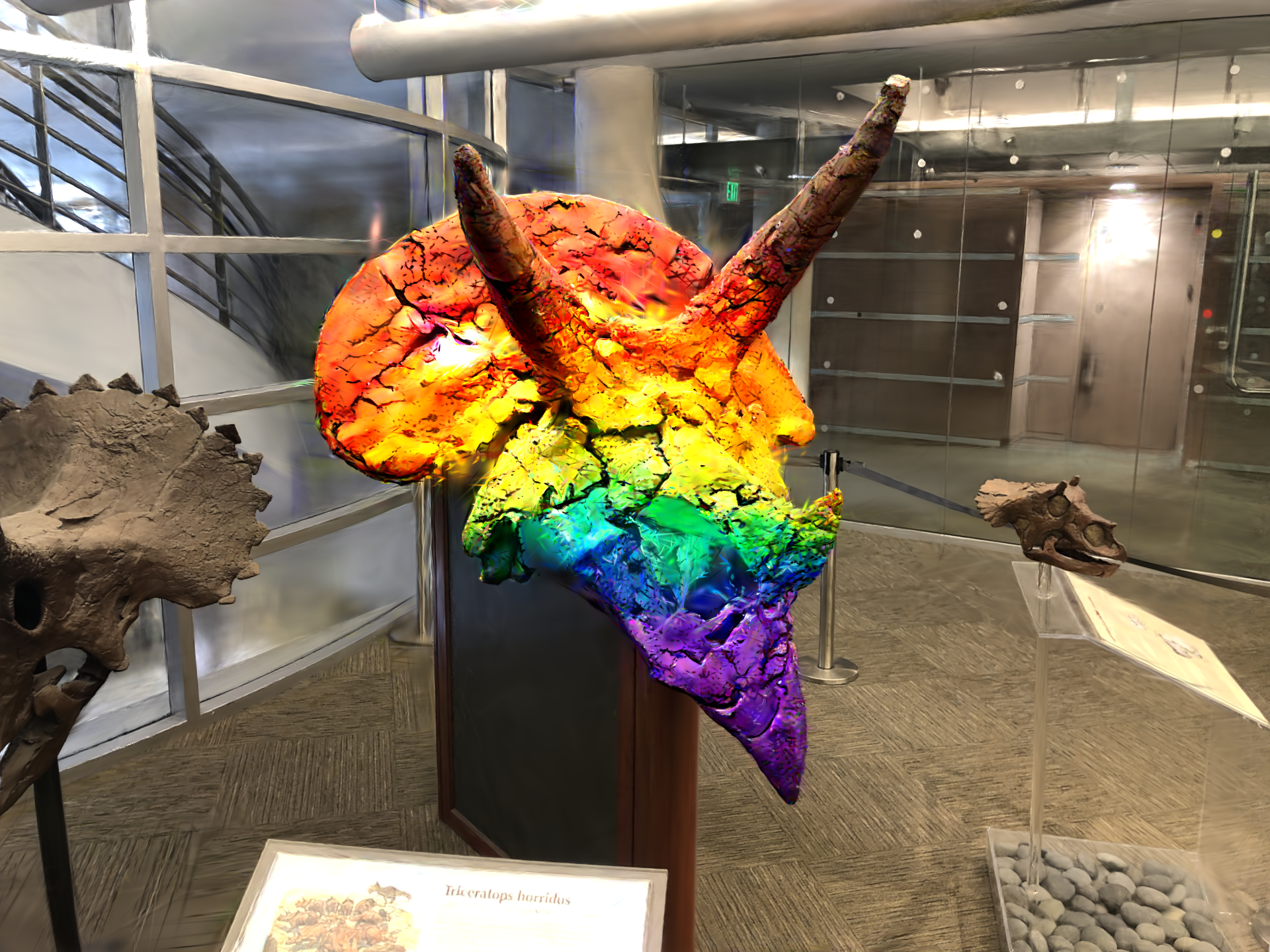}%
    \end{minipage}

    \vspace{5pt}
    \begin{minipage}[c]{0.95\linewidth}
        \centering
        \small
        \text{\textit{A tree stump with some leaves \textcolor{blue}{on fire}}}
    \end{minipage}%

    \vspace{1pt}
    \begin{minipage}[c]{0.99\linewidth}
        \includegraphics[width=0.19\linewidth]{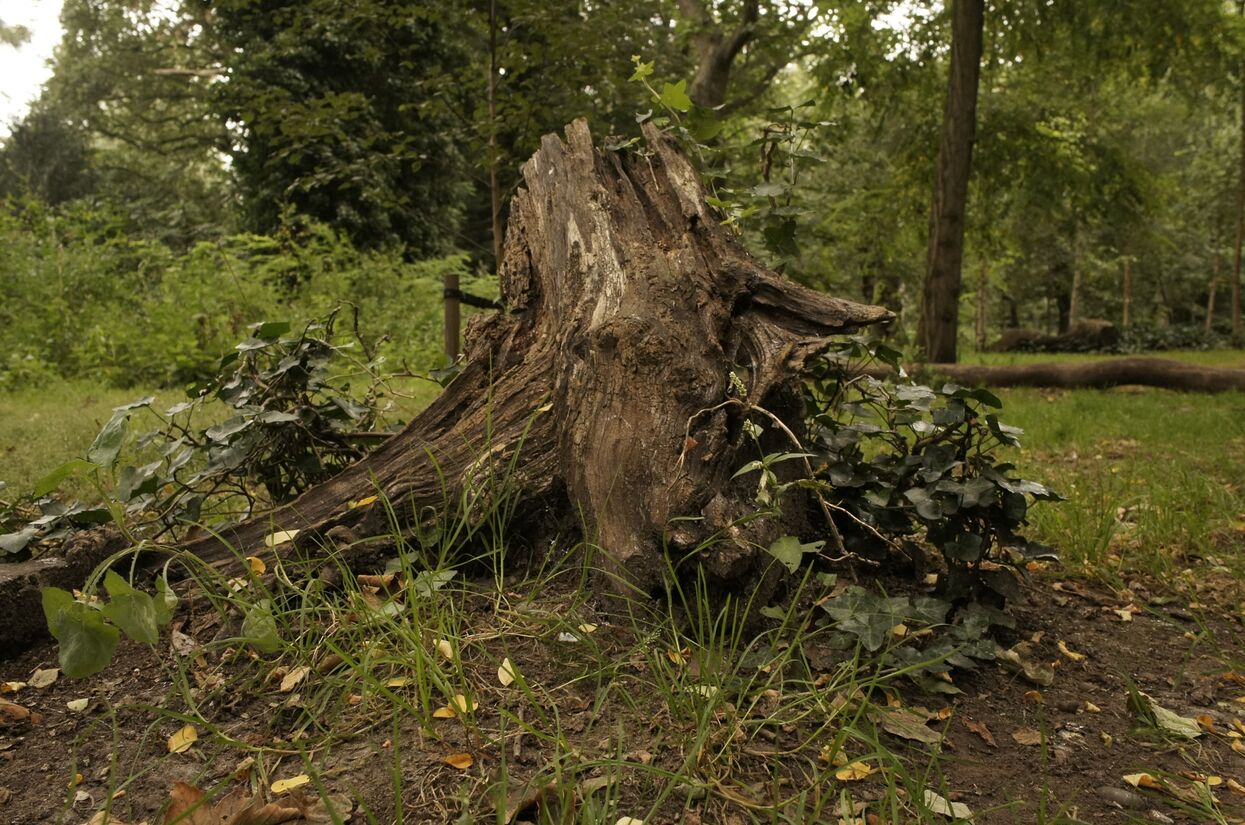}%
        \hspace{0.01mm}
        \includegraphics[width=0.19\linewidth]{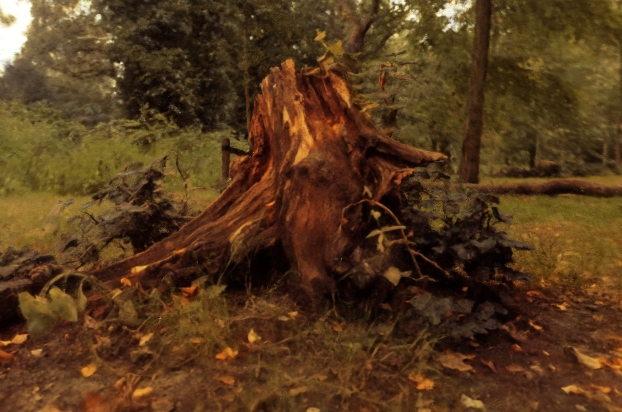}%
        \hspace{0.01mm}
        \includegraphics[width=0.19\linewidth]{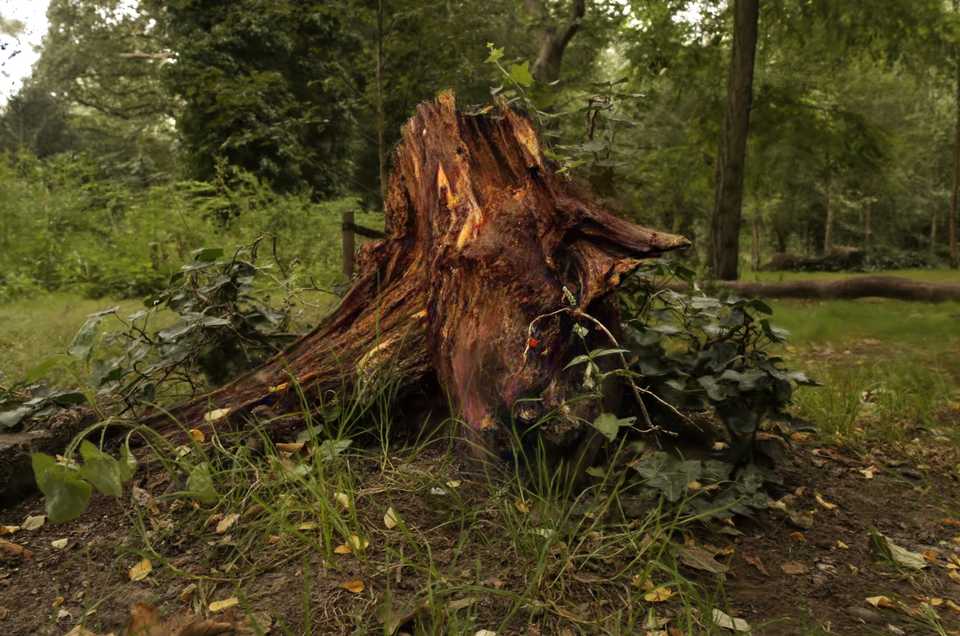}%
        \hspace{0.01mm}
        \includegraphics[width=0.19\linewidth]{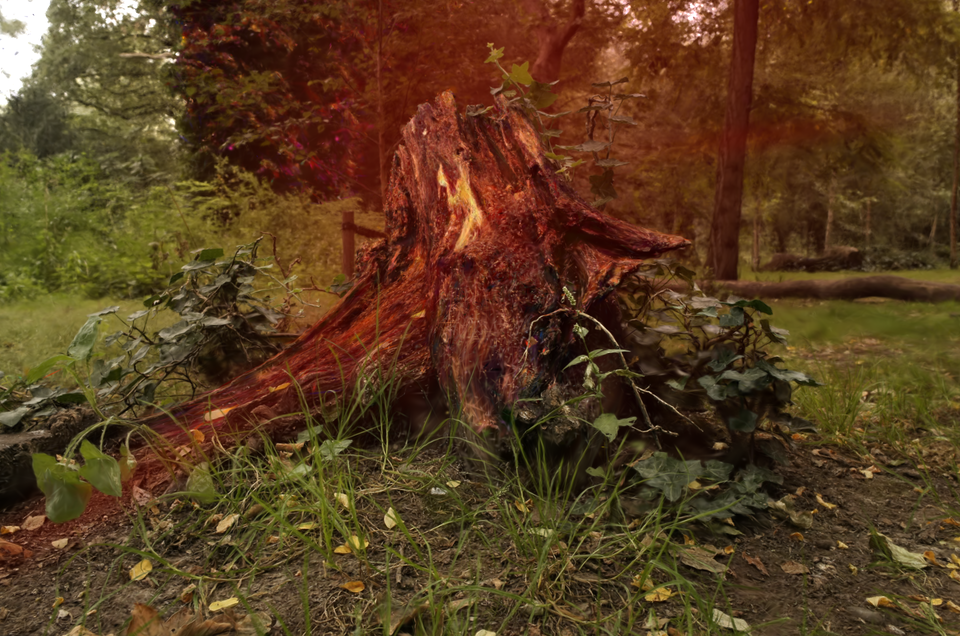}%
        \hspace{0.01mm}
        \includegraphics[width=0.19\linewidth]{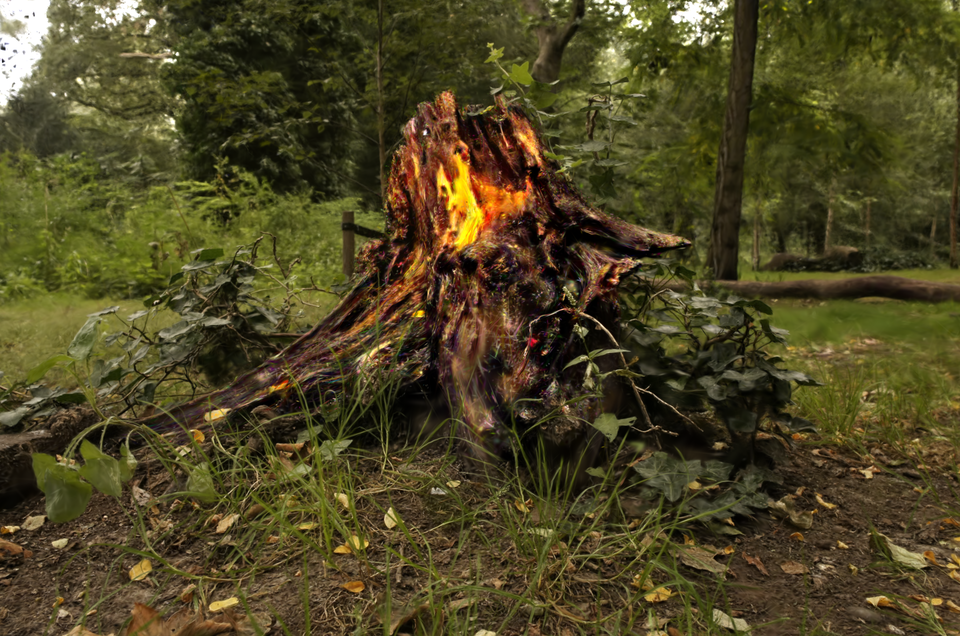}%
    \end{minipage}

    \vspace{1mm} 
    \begin{minipage}[c]{0.99\linewidth}
        \centering
        \small 
        \begin{minipage}{0.19\linewidth}
            \centering
            \raisebox{-0.5\height}{Input Views}
        \end{minipage}%
        \begin{minipage}{0.19\linewidth}
            \centering
            \raisebox{-0.5\height}{InstructN2N}
        \end{minipage}%
        \begin{minipage}{0.19\linewidth}
            \centering
            \raisebox{-0.5\height}{GaussianEditor}
        \end{minipage}%
        \begin{minipage}{0.19\linewidth}
            \centering
            \raisebox{-0.5\height}{DGE}
        \end{minipage}%
        \begin{minipage}{0.19\linewidth}
            \centering
            \raisebox{-0.5\height}{Ours}
        \end{minipage}%
    \end{minipage}

    \vspace{-1mm}
    \caption{Qualitative comparisons with related works. SDS demonstrates outstanding performance in effectively preserve source structure in the modified region.}
    \vspace{-5mm}
    \label{fig:com_all}
\end{figure*}

\noindent \textbf{Results.} We begin by evaluating our method, starting with a qualitative assessment. In Fig.~\ref{fig:com_all}, we present a comparison of results with competing methods. Our approach generates more visually appealing images that are better aligned with the editing instructions. In contrast, methods based on the Iterative Dataset Update (IDU) strategy, such as IN2N~\cite{instructnerf2023} and GS-Editor~\cite{chen2024gaussianeditor}, fail to produce the desired editing outcomes, resulting in blurrier or lower-fidelity reconstructions and noticeable artifacts. For example, in the scene of ``\textit{Spider-Man with a mask}", IN2N generates a mask with reduced fidelity, while GS-Editor produces a low-detail mask. In the multi-view consistency setup, DGE~\cite{chen2024dge} performs well on common attributes but is constrained to "rainbow" editing and tends to generate artifacts outside the segmentation mask. Our method works seamlessly with masks, producing results with rich details. 

In Tab.~\ref{tab:clip_metrics}, We present a quantitative comparison. Our method outperforms the baselines in terms of CLIP Similarity and CLIP Directional Similarity. Notably, Dire Sim is not sensitive with the editing quality, much focus on the instruction attributes. We conducted a user study with a survey of 55 participants to evaluate the editing quality. The results show that our method received the most popular votes.

\begin{table}[t]
    \centering
    \caption{Quantitative evaluations under 3D editing scenes.}
    \vspace{-3mm}
    \small
    \setlength{\tabcolsep}{0.5mm}{  
    \begin{tabular}{lccc}
        \hline
        \multirow{1}{*}{\textbf{Method}} & \multicolumn{1}{c}{\textbf{CLIP Sim} $\uparrow$} & \multicolumn{1}{c}{\textbf{Sim Dire} $\uparrow$} & \multicolumn{1}{c}{\textbf{User Study} $\uparrow$} \\
        \hline
        IN2N            & 0.1676          & 0.0707 & 14.54\%  \\
        DDS               & 0.1780          & 0.0401 & 5.45\%   \\
        GS-Editor  & 0.1758          & 0.0429 & 14.54\%  \\
        DGE                  & 0.1758          & 0.0563 & 23.63\%  \\
        \hline
        Ours                                     & \textbf{0.1846} & \textbf{0.0773} &  \textbf{41.81\%} \\
        \hline
    \end{tabular}}
    \vspace{-3mm}
    \label{tab:clip_metrics}
\end{table}

\subsection{2D Image Editing}
\label{subsec:exp-img}

\noindent \textbf{Dataset.} To evaluate the effectiveness of our method, we conduct experiments on the PIE-Bench dataset proposed by PNPInv~\cite{ju2024pnp}, which consists of 700 images with 9 editing types. Each image is annotated with the source and target prompts. 

\noindent \textbf{Baselines.} We compare our method with several classical editing methods based on DDIM~\cite{song2021denoising} inversion, including P2P~\cite{hertzprompt}, PNP~\cite{Tumanyan_2023_CVPR} and MasaCtrl~\cite{cao2023masactrl}. For optimization-based editing method, we compare with NT~\cite{mokady2023null} and StyleD~\cite{li2023stylediffusion}. Besides, we report the comparison with DT~\cite{ju2024pnp}. Further, we compare with DDS~\cite{hertz2023delta} and its extended method CDS~\cite{nam2024contrastive}. 

\noindent \textbf{Evaluation Metrics.} We follow DT~\cite{ju2024pnp} which uses several metrics to evaluate our method. We use the Structure Distance assessed by DINO score~\cite{caron2021emerging} to evaluate the structure distance between original and edited images. We also introduce several metrics to evaluating the background preservation, which includes LPIPS~\cite{zhang2018perceptual} and MSE. Besides, we introduce CLIP Similarity~\cite{wu2021godiva} to evaluate the text-image consistency between edited images and corresponding target editing text prompts. 

\begin{table}[t]
    \centering
    \caption{Quantitative evaluation in PIE-Bench dataset.}
    \vspace{-3mm}
    \small
    \setlength{\tabcolsep}{0.005mm}{       
    \begin{tabular}{lcccc}
        \hline
        \textbf{Method} & \textbf{Distance}$_{^{\times 10^3}}$ $\downarrow$ & \textbf{LPIPS}$_{^{\times 10^3}}$ $\downarrow$ & \textbf{MSE}$_{^{\times 10^4}}$ $\downarrow$ & \textbf{CLIP} $\uparrow$ \\
        \hline
        DDIM + P2P & 69.43 & 208.80 & 219.88 & 25.01 \\
        DDIM + PNP & 28.22 & 113.46 & 83.64 & 25.41 \\
        DDIM + MasaCtrl & 28.38 & 106.62 & 86.97 & 23.96 \\
        \hline
        NT + P2P & 13.44 & 60.67 & 35.86 & 24.75 \\
        StyleD + P2P & 11.65 & 66.10 & 38.63 & 24.78 \\
        DT + P2P & 11.65 & 54.55 & 32.86 & 25.02 \\
        \hline
        DDS & 14.74 & 50.58 & 45.09 & \underline{25.86} \\
        DDS + CDS & \underline{7.15} & \underline{33.14} & \underline{25.29} & 24.96 \\
        \hline
        Ours & 28.13 & 82.43 & 86.64 & \textbf{26.94} \\
        Ours + CDS & \textbf{6.90} & \textbf{32.15} & \textbf{24.21} & 25.12 \\
        \hline
    \end{tabular}}
    \vspace{-5mm}
    \label{tab:2d_metrics}
\end{table}

\begin{figure}[t]
    \centering
    \vspace{3pt}
    \begin{minipage}{1.0\linewidth}
        \centering
        \footnotesize
        \textnormal{\textit{A blue and white  \textcolor{blue}{bird} $\rightarrow$ \textcolor{blue}{butterfly} sits on a branch}}
    \end{minipage}

    \vspace{1pt}
    \begin{minipage}{1.0\linewidth}
        \centering
        \includegraphics[width=0.19\linewidth]{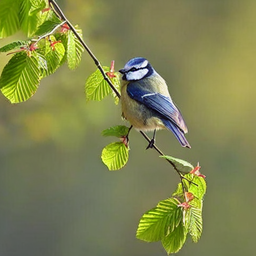}
        \includegraphics[width=0.19\linewidth]{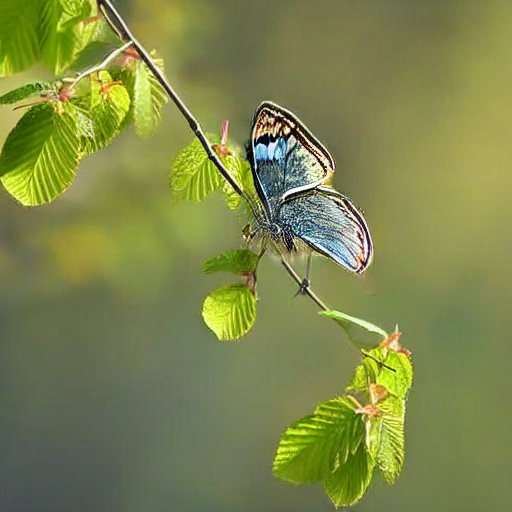}
        \includegraphics[width=0.19\linewidth]{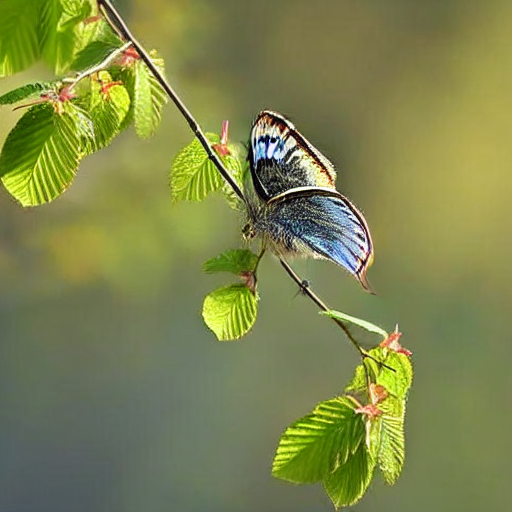}
        \includegraphics[width=0.19\linewidth]{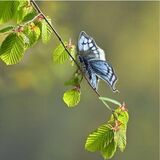}
        \includegraphics[width=0.19\linewidth]{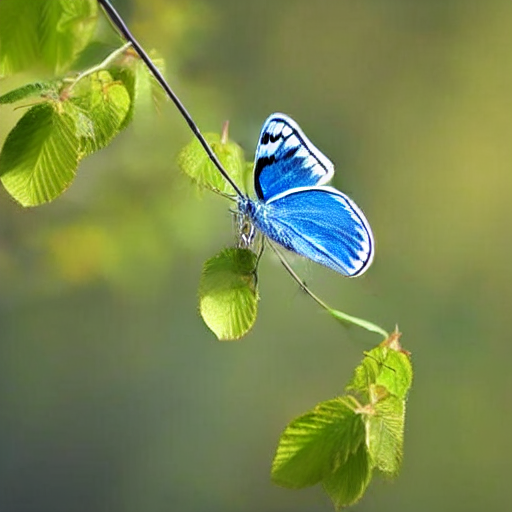}
    \end{minipage}

    \vspace{5pt}
    \begin{minipage}{1.0\linewidth}
        \centering
        \footnotesize
        \textnormal{\textit{\textcolor{blue}{Kids crayon drawing of} a man with a long beard and a long sword}}
    \end{minipage}

    \vspace{1pt}
    \begin{minipage}{1.0\linewidth}
        \centering
        \includegraphics[width=0.19\linewidth]{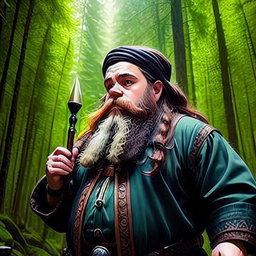}
        \includegraphics[width=0.19\linewidth]{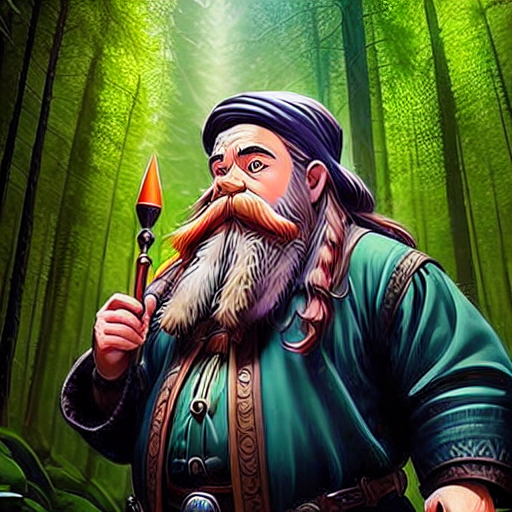}
        \includegraphics[width=0.19\linewidth]{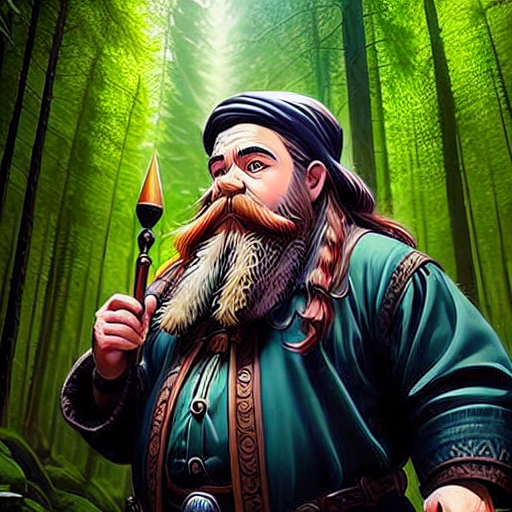}
        \includegraphics[width=0.19\linewidth]{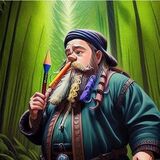}
        \includegraphics[width=0.19\linewidth]{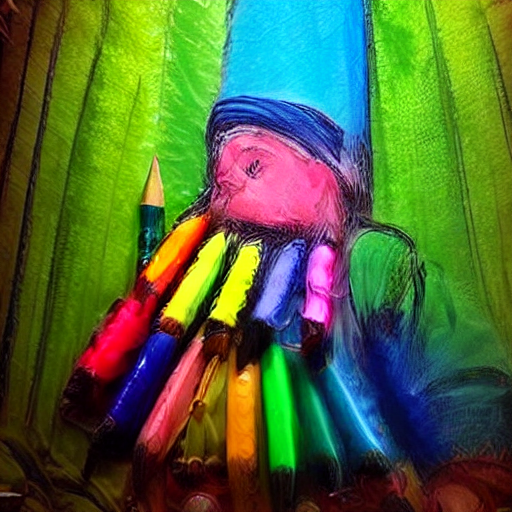}
    \end{minipage}

    \vspace{5pt}
    \begin{minipage}{1.0\linewidth}
        \centering
        \footnotesize
        \textnormal{\textit{\textcolor{blue}{Black and white sketch of} a young girl with painted hands and face}}
    \end{minipage}

    \vspace{1pt}
    \begin{minipage}{1.0\linewidth}
        \centering
        \includegraphics[width=0.19\linewidth]{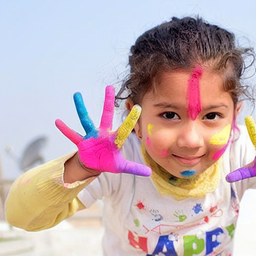}
        \includegraphics[width=0.19\linewidth]{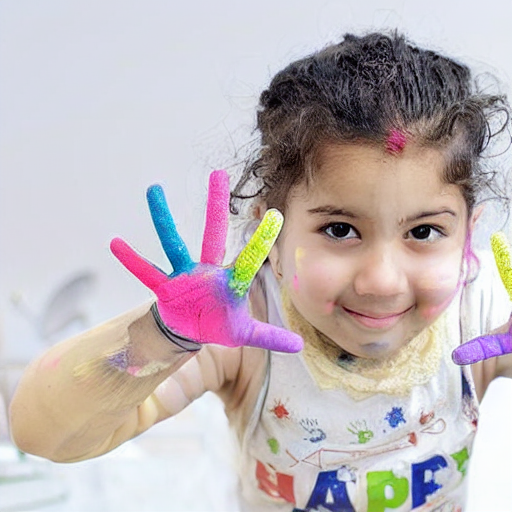}
        \includegraphics[width=0.19\linewidth]{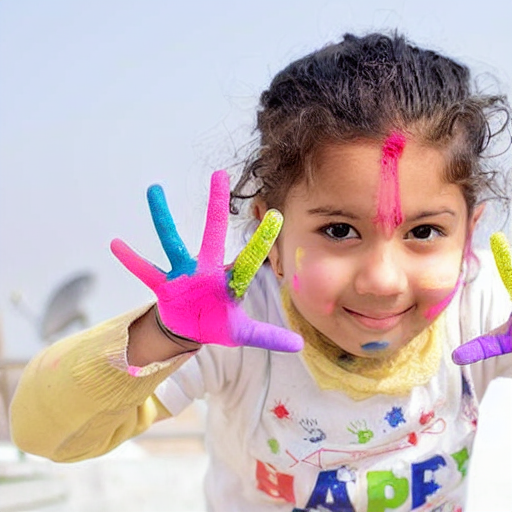}
        \includegraphics[width=0.19\linewidth]{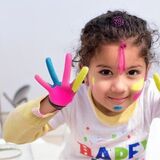}
        \includegraphics[width=0.19\linewidth]{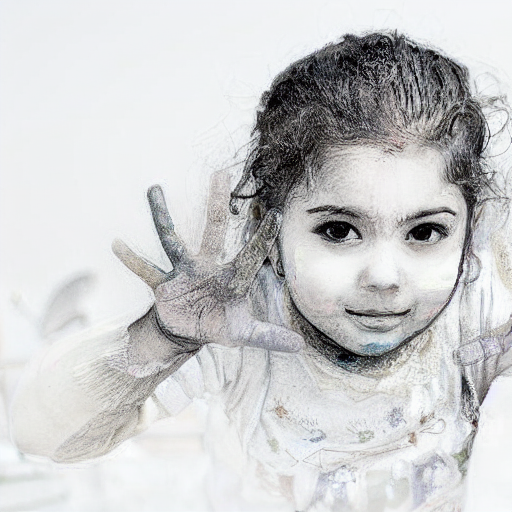}
    \end{minipage}

    \vspace{5pt}
    \begin{minipage}{1.0\linewidth}
        \centering
        \footnotesize
        \textnormal{\textit{A \textcolor{blue}{monkey} $\rightarrow$ \textcolor{blue}{man} wearing colorful goggles and a colorful scarf}}
    \end{minipage}

    \vspace{1pt}
    \begin{minipage}{1.0\linewidth}
        \centering
        \includegraphics[width=0.19\linewidth]{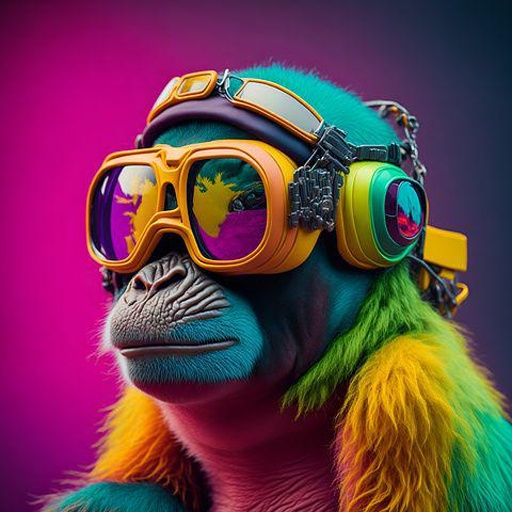}
        \includegraphics[width=0.19\linewidth]{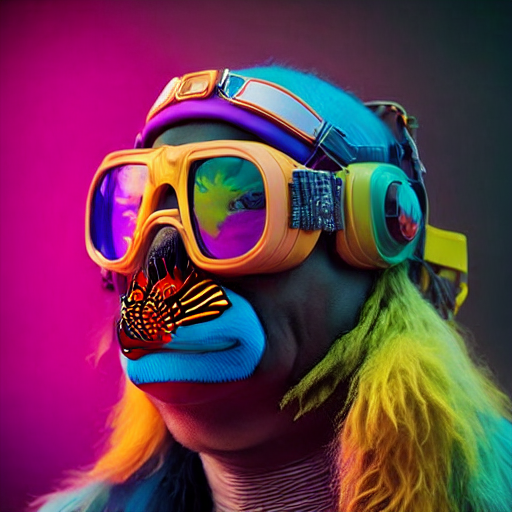}
        \includegraphics[width=0.19\linewidth]{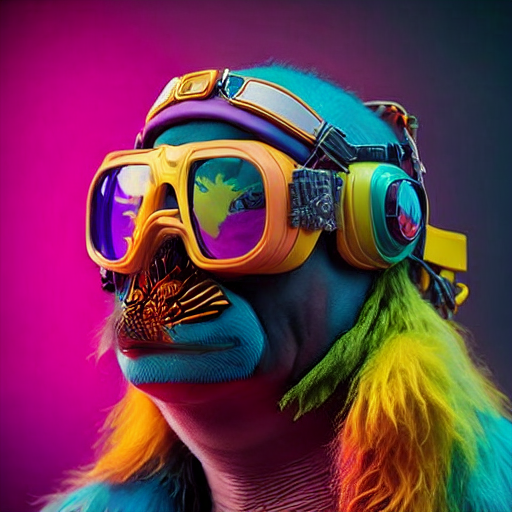}
        \includegraphics[width=0.19\linewidth]{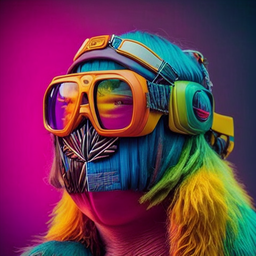}
        \includegraphics[width=0.19\linewidth]{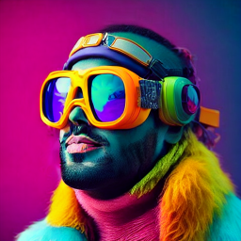}
    \end{minipage}

    \vspace{3pt}
    \begin{minipage}{1.0\linewidth}
        \centering
        \small
        Source \hspace{20pt} DDS \hspace{20pt} CDS \hspace{20pt} NT+P2P \hspace{20pt} Ours
    \end{minipage}
    \caption{Comparison of different editing methods on various objects and styles.}
    \vspace{-5mm}
    \label{fig:comp_2d}
\end{figure}

\noindent \textbf{Results.} We present a qualitative comparison of our method with competitors in Fig.~\ref{fig:comp_2d}. Our method generates images that are more aligned with the target prompts and preserve the source structure. In ``\textit{blue butterfly}'', ours successfully changes the color of the butterfly to blue, while DDS~\cite{hertz2023delta} and CDS~\cite{nam2024contrastive} generate similar color from source. Especially, ours method successfully changes the style of the image and generates appealing results, which is challenging for DDS-based methods. Compared to optimization-based methods, NT~\cite{mokady2023null} preserves the general source structure during the inversion process, but tends to discard some content as evident in the distortion of the girl's fingers in Fig.~\ref{fig:comp_2d}. Additionally, due to limitations in the editing methods, the editing results are unsuccessful.

In Tab.~\ref{tab:2d_metrics}, we present a quantitative comparison. Our method strikes a balance between structure distance and editability. Notably, we observe that the distance is much lower when no editing occurs, which is particularly visible in DDS-based methods applied to style editing. Our methods achieving better editing score but with a slightly higher structure distance. In terms of precise structure preservation, when combined with CDS~\cite{nam2024contrastive}, it achieves good preservation of the un-edited areas. Our full model achieves the best performance in the CLIP Similarity metric, demonstrating the effectiveness of our prompt enhancement branch. While CDS excels in preserving unedited regions, it suffers from the inferior editability of DDS-based methods. Optimization-based methods~\cite{mokady2023null, li2023stylediffusion} refine the inversion process, achieving excellent performance in structure preservation. However, they struggle with editing methods (like P2P), resulting in limited editability.

\subsection{Ablation Studies}
\label{subsec:exp-ab}

In this section, we conduct an ablation experiment to analyze different choices in our SSD. Due to space limitations, we first present a qualitative evaluation in the main text. Please refer to the Supp. for quantitative evaluation.

\noindent \textbf{The effectiveness of cross-trajectory.} In Sec.~\ref{subsec:ssd}, we have analyzed the necessity of cross-trajectory. This term make the optimization process more stable and provide the source content regularization in ours design, which is also the key difference between ours and Classifier Score Distillation(CSD)~\cite{yu2024texttod}. In Fig.~\ref{fig:teaser} and Fig.~\ref{fig:scaling_terms}, we present the comparison of the results with and without cross-trajectory. The results show that the cross-trajectory term can provide the direction of generating high-quality images. Please refer to the supplement for more details.

\noindent \textbf{The effectiveness of prompt-enhancement.} The enhancement of the target prompt branch is another key component in our method, which is designed to improve the editability aligned with the target prompt in 2D-image task. In Fig.~\ref{fig:comp_2d}, we observe a clear distinction from DDS in style editing. The results show that the prompt-enhancement term effectively overcomes the challenging from style editing.

\begin{figure}[t!]
    \centering
    \begin{minipage}{0.31\linewidth}
        \centering
        \includegraphics[width=\linewidth]{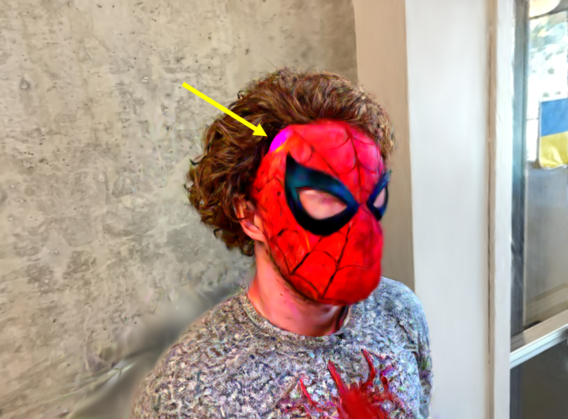}
        \vspace{-6mm} 
        \caption*{\small w/o ID Regular}
    \end{minipage}%
    \hspace{0.01\linewidth} 
    \begin{minipage}{0.31\linewidth}
        \centering
        \includegraphics[width=\linewidth]{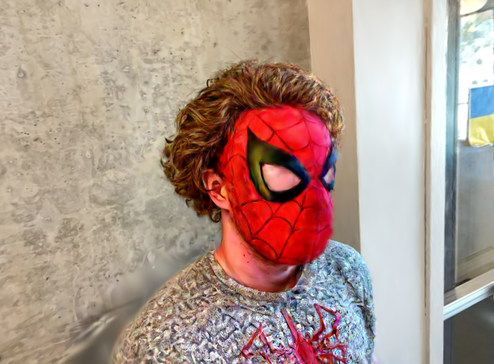}
        \vspace{-6mm} 
        \caption*{\small ID Regular $\times$ 1.0}
    \end{minipage}%
    \hspace{0.01\linewidth}
    \begin{minipage}{0.31\linewidth}
        \centering
        \includegraphics[width=\linewidth]{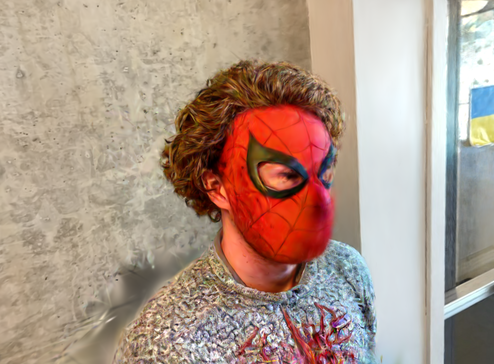}
        \vspace{-6mm} 
        \caption*{\small ID Regular $\times$ 2.0}
    \end{minipage}
    \caption{Effect of source latent regularization. In most experiments, the source ID term helps prevent partial gradient explosion. In the left image, the yellow arrow highlights an irregular color. As the weight of the ID term increases, the color becomes more regular, however, the spider on the character's chest is affected.}
    \vspace{-5mm}
    \label{fig:ab_id}
\end{figure}

\noindent \textbf{The effectiveness of ID regularization.} ID regularization is designed to ensure stable optimization in 3DGS. In Fig.~\ref{fig:ab_id}, we compare results with and without ID regularization. The area marked by the yellow arrow highlights its effect in 3D scene editing. However, excessive ID regularization may constrain editing quality by limiting certain attributes, presenting a trade-off in our design.

\section{Conclusions, Limitations, and Future Work}

In this work, we propose a novel method for text-guided image editing, capable of handling both 3D scenes and 2D images. Our approach is built on a score distillation framework that leverages the powerful priors of diffusion models. For editing tasks, we design an effective optimization strategy that produces high-quality results aligned with target prompts while ensuring stable and consistent optimization.

Our method achieves state-of-the-art performance in both 3D scene and 2D image editing, delivering realistic edits with excellent preservation of the original content. It demonstrates strong adaptability to various editing tasks and target prompts, making it a robust solution for complex scenarios. However, while effective, the optimization process is relatively time-intensive compared to recent one-step methods~\cite{xu2023infedit} or few-step approaches~\cite{deutch2024turboedit}. Future work could explore integrating advanced techniques such as LCM~\cite{luo2023latent} or SD-turbo~\cite{sauer2025adversarial}, which show potential for accelerating the optimization process~\cite{tian2024postedit}.

\small{\textbf{Acknowledgment:} This project is supported by the National Natural Science Foundation of China (62125201, U24B20174).}

{\small
\bibliographystyle{ieeenat_fullname}
\bibliography{./egbib}
}

\end{document}